\pdfoutput=1
\documentclass{article}

\usepackage{PRIMEarxiv}

\usepackage{url}
\usepackage{multirow}
\usepackage{subcaption}
\usepackage{listings}
\usepackage{xcolor}  
\usepackage{amssymb}
\usepackage{algorithm}
\usepackage{algorithmic}
\usepackage{comment}
\usepackage{amsmath}

\usepackage[utf8]{inputenc} 
\usepackage[T1]{fontenc}    
\usepackage{hyperref}       
\usepackage{url}            
\usepackage{booktabs}       
\usepackage{amsfonts}       
\usepackage{nicefrac}       
\usepackage{microtype}      
\usepackage{lipsum}
\usepackage{fancyhdr}       
\usepackage{graphicx}       
\AtBeginDocument{%
  }

\title{AutoGeTS: Knowledge-based Automated Generation of Text Synthetics for Improving Text Classification}

\author{
  Chenhao Xue \\
  Department of Engineering Science \\
  University of Oxford \\
  UK \\
  \texttt{chenhao.xue@eng.ox.ac.uk} \\
  \And
  Yuanzhe Jin \\
  Department of Engineering Science \\
  University of Oxford \\
  UK \\
  \texttt{yuanzhe.jin@eng.ox.ac.uk} \\
  \And
  Adrian Carrasco-Revilla \\
  FabLab, Inetum \\
  Madrid \\
  Spain \\
  \texttt{adrian.carrasco@inetum.com} \\
  \And
  Joyraj Chakraborty \\
  Department of Engineering Science \\
  University of Oxford \\
  UK \\
  \texttt{joyraj.chakraborty@eng.ox.ac.uk} \\
  \And
  Min Chen \\
  Department of Engineering Science \\
  University of Oxford \\
  UK \\
  \texttt{min.chen@eng.ox.ac.uk} \\
}

\date{July 22, 2025}

\begin{document}
\maketitle

\begin{abstract}
When developing text classification models for real world applications, one major challenge is the difficulty to collect sufficient data for all text classes. In this work, we address this challenge by utilizing large language models (LLMs) to generate synthetic data and using such data to improve the performance of the models without waiting for more real data to be collected and labelled. As an LLM generates different synthetic data in response to different input examples, we formulate an automated workflow, which searches for input examples that lead to more ``effective'' synthetic data for improving the model concerned. We study three search strategies with an extensive set of experiments, and use experiment results to inform an ensemble algorithm that selects a search strategy according to the characteristics of a class. Our further experiments demonstrate that this ensemble approach is more effective than each individual strategy in our automated workflow for improving classification models using LLMs.
\end{abstract}


\maketitle

\section{Introduction}\label{sec:Introduction}
In industrial applications of text classification the classes are typically defined according to semantic grouping as well as organizational function. Critical impediments to developing robust text classification models for such applications include (i) noticeably imbalanced class sizes, data scarcity in some classes, and (ii) changes of the categorization scheme due to organizational changes. One group of examples are automated ticketing systems in different companies and organizations for processing users' messages and distributing them to different services, e.g., IT issues, building problems, operational incidents, and service requests \cite{al2021machine}.

\begin{figure*}[htpb]
\begin{center}
\includegraphics[width=0.8\linewidth]{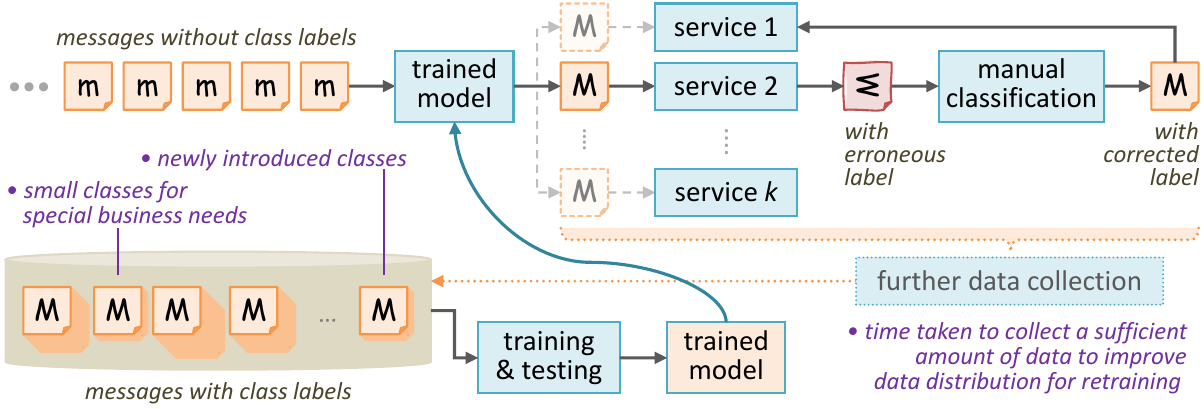}
\end{center}
\caption{The workflow for developing and deploying a classification model in an industrial ticketing system, and the main obstacles impacting on the performance of the model.}
\label{fig:Challenges}
\end{figure*}

As illustrated in Figure \ref{fig:Challenges}, a model is initially trained on a set of labeled tickets. Classification errors necessitate manual classification and redistribution, incurring delays and costs \cite{li2022survey}. With the gradual changes within each organization, messages of certain semantics may become less frequent, while those with other semantics (including new semantics) become more frequent. Over time, the model performance degrades \cite{Gandla_2024}. On the one hand, such a model needs to be improved regularly. On the other hand, identifying worsened performance of a model is usually not accompanied by adequate training data for improving the model.

Synthetic data has been used to overcome the limitations of real-world data, addressing data scarcity, sensitivity, or collection cost \cite{lu2023machine,patki2016synthetic} in many fields, such as computer vision and NLP \cite{mumuni2024survey}, medical imaging \cite{frid2018synthetic}, autonomous driving systems \cite{song2023synthetic}, finance \cite{chale2022generating} and cybersecurity \cite{potluru2023synthetic}. In this work, we focus on providing a novel and cost-effective solution to improve models for classifying short messages in ticketing systems.

In particular, we conducted a large number of experiments to study the approaches for finding message examples for LLMs to generate synthetic data for improving models. Building on our analysis of the experimental results, we outlined a novel algorithm for controlling the model improvement workflow by distributing the computational resources (e.g., for searching examples, generating synthetic data, retraining models, and testing models) according the knowledge gained from systematic exploration of the algorithmic space for identifying effective message examples. We also validated out findings using other datasets and data synthesis method.     

\section{Related Work}\label{sec:RelatedWork}
Synthetic data has increasingly been used to assist in various data science tasks across many domains \cite{meier1988testing,bersano1997synthetic}. Bootstrapping \cite{efron1992bootstrap,breiman1996bagging} is one of the early \emph{data synthesis} methods. It resamples original data to simulate desired distributions and improve model performance \cite{sutton2005classification}. To overcome the shortcomings of imbalanced datasets, synthetic data was used in conjunction with the common method to over-sample the minority classes and under-sample the majority classes \cite{chawla2002smote}. Today, \emph{data augmentation} encompasses a family of techniques that transform existing data in a dataset to generate synthetic data with desired properties, such as increasing diversity, changing distributions, filling in missing data, and so on \cite{jaderberg2014synthetic}. 

Machine learned models, such as Generative Adversarial Networks (GANs) \cite{goodfellow2014generative}, were used to generate synthetic data with a high level of realism and complexity. Studies have shown that models trained on GAN-generated data often can perform comparably to those trained on real data \cite{zhang2017adversarial, cortes2020analysis}. For example,
Frid-Adar et al. \cite{frid2018gan} enhanced liver lesion diagnosis using GAN-generated images.
Yale et al. \cite{yale2020generation} demonstrated comparable performance using GAN-generated electronic health records for ICU patient predictions.
Croce et al. \cite{croce2020gan} demonstrated their effectiveness in generating realistic text for NLP tasks.
He et al. \cite{he2022generate} explored task-specific text generation. However, GAN-generated data for text classification often lacks semantic coherence and relevance to specific tasks \cite{torres2018generation}. 

Recent advancements in large language models (LLMs), such as GPT-2 \cite{radford2019language}, provide new approaches to overcome these limitations. LLMs excel in few-shot and zero-shot learning \cite{brown2020language, wang2021towards}, adapting to unseen tasks and generating contextually relevant data that improves model robustness. Yoo et al.'s GPT-3Mix \cite{yoo2021gpt3mix} demonstrated LLMs' capability to generate diverse, effective synthetic data for text classification through careful \emph{prompt engineering}. Prompt optimization strategies have shown that carefully crafting input prompts can significantly impact the quality of generated data \cite{wang2023promptagent}. Automated search techniques for identifying optimal prompts, such as those used in AutoPrompt \cite{shin2020autoprompt, xu2024knowledge}, offer a potential solution to improve synthetic data generation. 

One aspect in prompt engineering is to select existing data objects as input prompts (examples). The effectiveness of such example data objects becomes crucial factor in synthetic data generation. Several methods have been proposed to enable the selection of effective input examples, ranging from uniform distribution to human selection with the aid of visualization \cite{li2023synthetic, Jin2024iGAiVAIG}. Motivated by an industrial application, this work aims to provide an automated technique that can perform example selection much faster than the human selection approach \cite{Jin2024iGAiVAIG}, while simulating some behaviors in human selection processes.

Beyond data synthesis and prompt engineering, ensemble methods has emerged as another promising approach for improving synthetic data generation. Xu et al. \cite{xu2022adaptive} proposed AdaDEM, which ensembles multiple convolutional networks at different granularity levels to optimize synthetic data selection for text classification. Similarly, Zhou et al. \cite{zhou2021metaaugment} introduced MetaAugment, an ensemble framework based on reinforcement learning that dynamically selects augmentation strategies per class, ensuring adaptive augmentation. Agbesi et al. \cite{agbesi2024mutcelm} developed MuTCELM, integrating multiple sub-classifiers to capture distinct linguistic features in an ensemble framework.

In this work, we extract useful knowledge in the testing data in our model improvement workflow AutoGeTS and use the knowledge to select an ensemble of effective methods for identifying example data.

\section{Methods}
\label{sec:Methods}
\begin{figure*}[ht]
\begin{center}
\includegraphics[width=\linewidth]{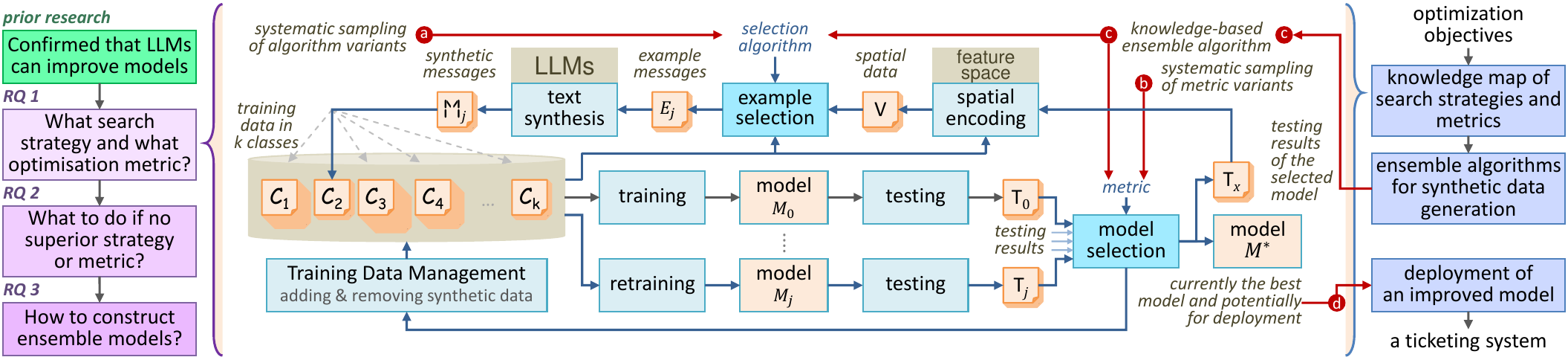}
\end{center}
\caption{AutoGeTS architecture: (left) the research questions (RQs) instigated its design. (middle) its experimental workflow used for answering research questions, creating a knowledge map, and enabling a model-improvement operation. (right) its operational pipeline for improving a model. It can support the development and maintenance of multiple ticketing systems.}
\label{fig:AutoGeTS}
\end{figure*}

As the prior work \cite{Jin2024iGAiVAIG} has already confirmed that synthetic data generated using LLMs can improve classification models used in automated ticketing systems, this work focuses on three research questions as shown in the upper part of Figure~\ref{fig:AutoGeTS}. Firstly, we conducted a large number of structured experiments to understand the effect of different variants of the algorithm for finding suitable message examples on the performance of model improvement. Ideally, we could discover a superior algorithm. We will describe our experimental method in this section and report the results and analysis in 
Section \ref{sec:Experiments}. Our analysis showed the absence of such a superior algorithm. This led to the second and third research questions, which will be discussed in Section \ref{sec:Ensemble}.

\vspace{1mm}\noindent
\textbf{Objectives for Model Optimization.}
Ticketing systems deployed in specific organizational environments often face different, sometimes conflicting, requirements. Typical business requirements and related performance metrics include:

\begin{itemize}
    \item[R1.] The accuracy of every class should be as high as possible and above a certain threshold. One may optimize a model with a performance metric such as class-based \emph{balanced accuracy} or \emph{F1-score} as the objective function, with each threshold value as a constraint.
    \item[R2.] The overall classification accuracy of a model should be as high as possible and above a certain threshold because misclassified messages lead to undesirable consequences. One may optimize a model with a global performance metric, such as overall \emph{balanced accuracy} and overall \emph{F1-score}.
    \item[R3.] The recall for some specific classes (e.g., important) should be as high as possible and above a certain threshold in order to minimize the delay due to the messages in such a class being sent to other services. Class-based \emph{recall} is the obvious metric for this requirement.  
\end{itemize}

These requirements inform the definition of objective functions and constraints for model optimization. However, as the use of LLMs to generate synthetic data to aid ML is a recent approach \cite{Jin2024iGAiVAIG}, it is necessary to understand how different example selection algorithms for LLMs may impact the performance of optimization with different metrics. This led to \textbf{\emph{research question 1}} in Figure \ref{fig:AutoGeTS}.%

\begin{figure*}[ht]
    \centering
    \begin{tabular}{@{}c@{\hspace{3mm}}c@{\hspace{3mm}}c@{}}
        \includegraphics[width=44mm]{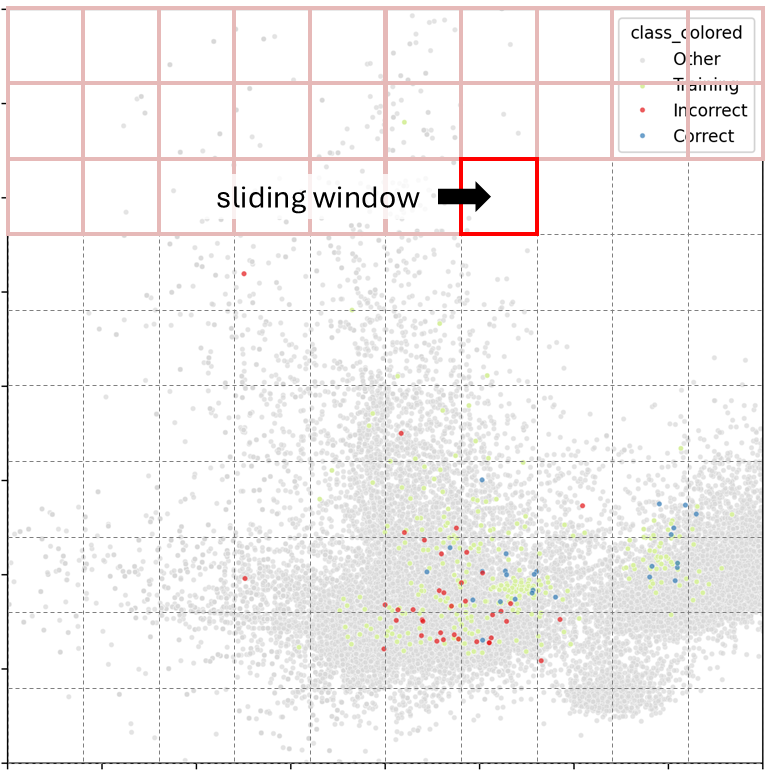} &
        \includegraphics[width=44mm]{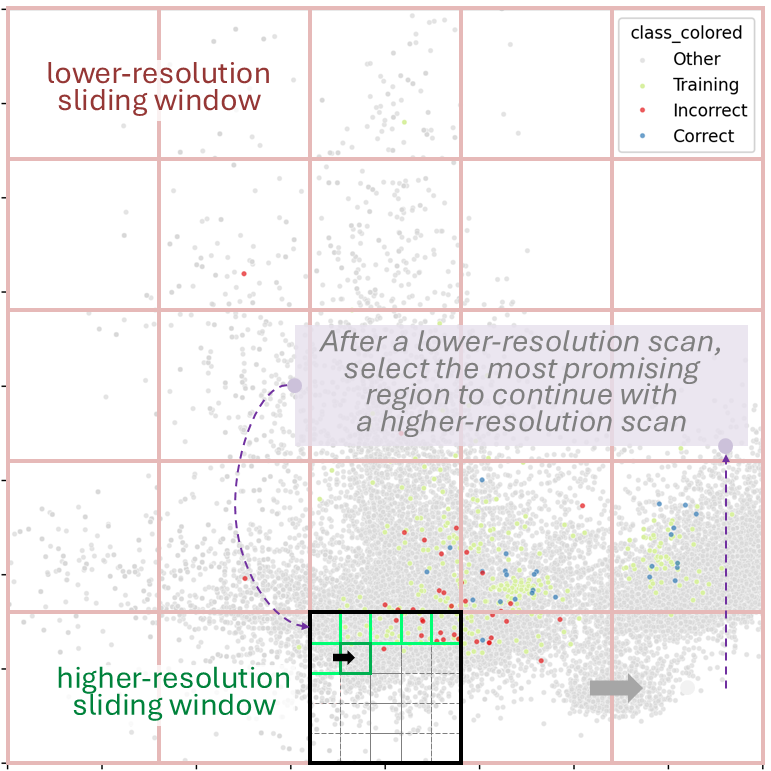} &
        \includegraphics[width=44mm]{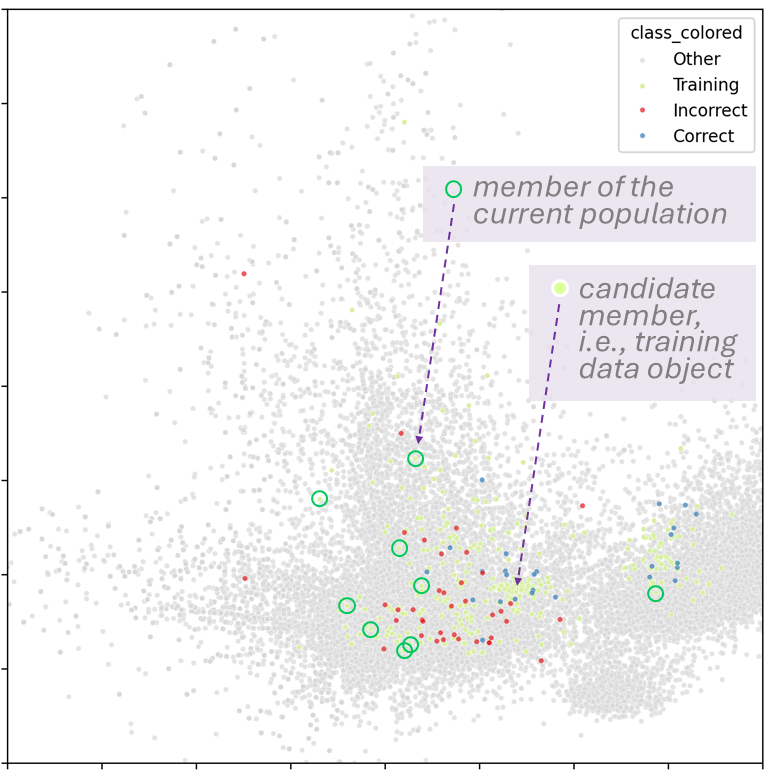} \\
        (a) sliding window & (b) hierarchical sliding window & (c) genetic algorithm
    \end{tabular}
    \caption{The three strategies experimented in this work for selecting example messages to be used by LLMs for generating synthetic data. The examples are only selected from training data of a particular class (light green dots), while blue, red, and gray dots illustrate contextual information, including correct and incorrect testing results and the data of other classes.}
    \label{fig:strategies}
\end{figure*}

\vspace{1mm}\noindent
\textbf{AutoGeTS Architecture and Workflow.}
The AutoGeTS architecture includes an experimental environment and an operational pipeline, which are illustrated in the middle and right parts of Figure~\ref{fig:AutoGeTS}. The experimental environment was initially used to answer \textbf{\emph{research question 1}}, and later to support the operational pipeline, which will be detailed in Section \ref{sec:Ensemble}.

Given a model $M_0$ to be improved and a set of improvement requirements (overall or class-specific), the improvement process is an optimization workflow as shown in the middle of Figure~\ref{fig:AutoGeTS}. To answer \textbf{\emph{research question 1}}, we systematically sample different variants of selection algorithm, run the workflow for each algorithmic sample, collect a large volume of results, and analyze the performance of these variants in obtaining the best model.

As reported in \cite{Jin2024iGAiVAIG}, using class-based visualization plots similar to those shown in Figure \ref{fig:strategies}, humans can select example messages from the training data and provide LLMs with these examples to generate synthetic data. To conduct experiments at a large scale, we replaced humans in the process. For a selected class $C$, visualization plots are automatically generated for every class. All training and testing data are mapped onto data points in an $n$-D feature space (in our work, $n=20$). Each plot depicts data points in two feature dimensions, with blue for correct testing data points in $C$, red for incorrect ones in $C$, light green for training data points in $C$, and gray for all data points in other classes. Note that an algorithm selects example messages only from the green data points.

An algorithm for selecting example messages can have many variations and their effectiveness depends on (1) the training datasets, (2) the structures and training parameters of models, (3) optimization metrics, and (4) the LLMs used and their control parameters. While it is not feasible to sample all algorithmic variations and their related conditions finely, our experiments covered a broad scope:%

\begin{enumerate}
    \item[0.] We compared three search strategies for example selection, namely \emph{Sliding Window} (SW), \emph{Hierarchical Sliding Window} (HSW), and \emph{Generic Algorithm} (GA). We conducted a pilot study to ensure that the parameters (e.g., window size, crossover method, etc.) of each strategy is fairly optimized.%
    \item[1.] In addition to a real world dataset collected in an industrial ticketing system, we also experimented with two other datasets, TREC-6 \cite{li2002learning} and Amazon Reviews 2023 \cite{hou2024bridging}.%
    \item[2.] We optimized the structure and training parameters for our baseline model $M_0$, and use the same structure and training parameters consistently throughout experiments for a training dataset.%
    \item[3.] We compared optimization performance using a number of metrics, including
     \emph{balanced accuracy}, \emph{F1-score}, and \emph{recall} for each of the 15 classes as well as for the overall dataset.%
    \item[4.] We used the GPT-3.5's API with a zero-shot prompt template. We conducted a pilot study to ensure that its control parameters were fairly optimized for examples selected using all three algorithms (i.e., SW, HSW, and GA). We compared the data generated using the GPT-3.5's API and the Easy Data Augmentation (EDA) tool \cite{wei2019eda}.%
\end{enumerate}

\noindent
\textbf{Strategies for Example Selection.}
Given $m$ text messages in a training dataset, there are a total of $2^m$ different combinations for selecting $1 \leq k \leq m$ messages as input examples for LLMs to generate synthetic data. Probabilistically, synthetic data generated using any of these $2^m$ combinations might help improve a model. However, testing all $2^n$ combinations falls into the NP category.  

The previous non-automatic work \cite{Jin2024iGAiVAIG} has found uses visual clusters of negative testing results (red dots) in different 2D feature spaces to guide the selection of examples from training data (light green dots) in an individual class. They found that the approach was relatively effective, though some attempts would fail to result in an improved model. Based on this finding, we consider three strategies for example selection:

$\blacktriangleright$ \emph{Sliding Window} (SW) --- As illustrated in Figure \ref{fig:strategies}(a), this simple automated strategy scans a 2D feature space square by square, and in each attempt, selects $k$ training data points randomly as examples. The total number of \emph{attempts} is more than the number of windows.

$\blacktriangleright$ \emph{Hierarchical Sliding Window} (HSW) --- In the non-automatic work \cite{Jin2024iGAiVAIG}, ML developers make intuitive decisions about the sizes of regions for selecting examples according to the visual patterns they are observing. In order to provide more flexibility than the SW approach, this automated strategy has a larger sliding window at level 1, and when encountering a promising or interesting window, it hierarchically makes more attempts with reduced window sizes as illustrated in Figure \ref{fig:strategies}(b). The specification of $k$ is the same as SW. The total number of \emph{attempts} is the sum of the attempts in every window examined at every level.
 
$\blacktriangleright$ \emph{Genetic Algorithm} (GA) --- In the non-automatic work \cite{Jin2024iGAiVAIG}, ML developers do not just rely on simple visual clusters of erroneous testing data points, and they sometimes select examples from a few separated regions. To enable more flexibility than the SW and HSW approaches, this automated strategy allows examples to be selected from any training data in a 2D feature space. Given $m_i$ training data points in a particular class $C_i$, the GA maintains a population of $r$ chromosomes, each of which has $1 \leq s \leq m_i$ genes that are switched on representing $s$ training data points selected as examples. In every iteration, GA evolves the population, while making attempts to improve the model with chromosomes that have not been previously tested. The total number of \emph{attempts} is the sum of the attempts in every iteration.

\vspace{1mm}\noindent
\textbf{Objective Functions for Optimization.}
As illustrated in Figure \ref{fig:AutoGeTS}, the above three strategies provide alternative algorithms to the \textbf{example selection} process, with which the workflow in the middle of the figure makes numerous \emph{attempts} to find different subsets of training data points of a particular class as examples for LLMs to generate synthetic data. The synthetic data is then added into the training data, and the classification is retrained and tested. Therefore, the workflow is essentially an optimization process to find a set of examples $E^*$ that results in a model $M^*$ that is considered to be the best model, i.e., given $\mathbb{E}$ consisting of all sets of examples attempted in a multi-iteration workflow, the best model $M^*$ is:  
\begin{align*}
    \text{Objective:}& \quad  \arg \max \bigl(f(M_0), \max_{E_j \in \mathbb{E}} f(M_j)\bigr)
    \quad j=1,2,\ldots,t \\
    \text{Subject to:}& \quad M_j \leftarrow \text{TRAIN} \bigl(M^\square, D_\text{tn}, \text{LLM}(E_j) \bigr) \quad j=1,2,\ldots,t\\
    & \quad f(M_j) \leftarrow \phi\bigl(\text{TEST}(M_j, D_\text{tt})\bigr) \quad j=1,2,\ldots,t
\end{align*}
where
$D_\text{tn}$ and $D_\text{tt}$ are the original training and testing data respectively,
$M_0$ is the baseline model to be improved,
$t = \| \mathbb{E} \|$ is the total number of attempts (i.e., sets of example data points selected from $D_\text{tn}$), 
$M_j (j \in [1, t])$ are sampled models in the optimization process,
TRAIN() and TEST() are the processes for training and testing a classification model $M_j$,
$f$ is the measurement of the testing results with a statistical measure $\phi$,
$M^\square$ is the structure of a classification model, and
$\text{LLM}()$ is the process for LLMs to generate synthetic data using a set of examples $E_j$.

In ticketing systems, \emph{recall} for an individual class $\phi^\text{cr}$ is a measure that client organizations pay attention to because improving $\phi^\text{cr}$ reduces the amount of manual classification in Figure \ref{fig:Challenges}. On the other hand, for a specific class, when we consider both TP and FP (true and false positive) messages, for a small class, the FP total (\#FP) can easily overwhelm the TP total (\#TP). Measures, such as \emph{accuracy} and \emph{precision}, can be overly biased by \#FP. 
For this reason, we focus on \emph{recall} $\phi^\text{cr}$ and \emph{balanced accuracy} $\phi^\text{cba}$ to measure the improvement made in the context of a specific class $C_i$. Meanwhile, to measure the overall performance of a model, we use overall \emph{balanced accuracy} $\phi^\text{oba}$ and overall F1-score $\phi^\text{of1}$. In summary:
\[
    \phi^\text{cr}_\text{c} = \frac{\text{\#TP}_c}{\text{\#TP}_c + \text{\#FN}_c} \quad
    \phi^\text{cba}_\text{c} = \frac{1}{2}\Bigl(\frac{\text{\#TP}_c}{\text{\#TP}_c + \text{\#FP}_c} + \frac{\text{\#TN}_c}{\text{\#TN}_c + \text{\#FN}_c}\Bigr)
\]
\[
    \phi^\text{of1}= \frac{2\text{\#TP}}{2\text{\#TP} + \text{\#FP} + \text{\#FN}} \quad
    \phi^\text{oba}= \frac{1}{2} \Bigl(\frac{\text{\#TP}}{\text{\#TP} + \text{\#FP}} + \frac{\text{\#TN}}{\text{\#TN} + \text{\#FN}} \Bigr)
\]
where the subscript $_c$ indicated the total values in a class $C$.


\section{Single-Phase Experiments and Results}
\label{sec:Experiments}
In this work, we first conducted single-phase experiments to answer \textbf{\emph{research question 1}} as discussed in Section \ref{sec:Methods}. Here ``single-phase'' means that a baseline model $M_0$ is improved by running the workflow using $\mathbb{E}_a$, i.e., examples selected from the training data of a single class $C_a$.
As indicated by (a) and (b) in Figure \ref{fig:AutoGeTS}, we systematically sampled three algorithmic variants and four metric variants, together with the variations of classes where example messages were selected.
In Section \ref{sec:Ensemble}, we will report that the systematic testing can provide an answer to \textbf{\emph{research questions 2 \& 3}}, while supporting multi-phase optimization, with which $M_0$ is improved by running the workflows in multiple phrases, using examples from different classes, i.e., $\mathbb{E}_a$, $\mathbb{E}_b$, etc.

For \textbf{\emph{research question 1}}, we sought answers in several aspects:
\begin{itemize}
    \item How does each of the three strategies (SW, HSW, GA) perform under different testing conditions (e.g., performance metric, time allowed for finding $M^*$, etc.).
    \item Is there a superior strategy for each metric $\phi$ as an objective function?
    \item How does optimization using one metric $\phi$ affect the model performance measured with other metrics?
    \item How does improvement made using examples from one class affect the overall performance of the model and its performance in other classes?  
    \item Is synthetic data generated by LLMs comparable with traditional data augmentation method? For this, we also conducted experiments with data generated using the Easy Data Augmentation (EDA) tool \cite{wei2019eda}.
    \item Are the findings obtained in our experiments data-dependent? For this, we also conducted experiments on two public datasets, TREC-6 \cite{li2002learning} and Amazon Reviews 2023 \cite{hou2024bridging}.
\end{itemize}

\noindent
\textbf{Experiment Setup.}
Our main experiments were conducted using a real-world dataset collected in an industrial ticketing system. It has 39,100 pieces of messages as training and testing data objects. Their labels fall into 15 classes for different services (Figure \ref{fig:Challenges}). As shown in Table \ref{tab:M0 Model}, the dataset is highly imbalanced, with
some classes have less than 1\% of the total data objects.
Because of the imbalance, we split the dataset into 60\% for training, 20\% for optimization testing, and 20\% for performance testing outside the optimization process.

Developing models with imbalanced data is a common phenomenon among almost all ticketing systems in different organizations.

\begin{table}[ht]
\caption{The performance of the original CatBoost model $M_0$}
\label{tab:M0 Model}
\centering
\renewcommand{\arraystretch}{1}
\begin{tabular}{@{}|c|c|c|c|c|@{}}
    \hline
    & & \textbf{Balanced} & & \\
    \textbf{Class} & \textbf{Class Size} & \textbf{Accuracy} & \textbf{Recall} & \textbf{F1-Score} \\
    \hline
    T1 & 8529 & 0.986 & 0.979 & 0.977 \\
    T2 & 11350 & 0.950 & 0.941 & 0.921 \\
    T3 & 4719 & 0.952 & 0.914 & 0.922 \\
    T4 & 1387 & 0.899 & 0.801 & 0.859 \\
    T5 & 2755 & 0.889 & 0.794 & 0.794 \\
    T6 & 1888 & 0.821 & 0.665 & 0.623 \\
    T7 & 1963 & 0.883 & 0.780 & 0.766 \\
    T8 & 1028 & 0.828 & 0.665 & 0.672 \\
    T9 & 1466 & 0.861 & 0.747 & 0.680 \\
    T10 & 1699 & 0.761 & 0.540 & 0.554 \\
    T11 & 471 & 0.973 & 0.947 & 0.967 \\
    T12 & 358 & 0.742 & 0.484 & 0.608 \\
    T13 & 180 & 0.666 & 0.333 & 0.469 \\
    T14 & 764 & 0.772 & 0.548 & 0.607 \\
    T15 & 543 & 0.726 & 0.452 & 0.596 \\
    \hline
    Overall & 39100 & 0.923 & 0.856 & 0.856 \\
    \hline
\end{tabular}
\end{table}

Referring to the workflow in the middle of Figure \ref{fig:AutoGeTS}, we used GPT-3.5 (version: 2023-03-15-preview) to generate synthetic text with main parameters \emph{temperature} = 0.7, \emph{max tokens} = 550, \emph{top p} = 0.5, \emph{frequency penalty} = 0.3, and \emph{presence penalty} = 0.0. The original model ($M_0$) was developed in an industrial application using CatBoost. For ensuring this research relevant to the industrial application, we trained all models using CatBoost consistently with fixed hyperparameters, i.e., \emph{number of iterations} = 300, \emph{learning rate} = 0.2, \emph{depth} = 8, \emph{L2 leaf regularization} = 1.

\begin{table*}[t]
    \centering
    \caption{(a) The results of systematic testing of 180 combinations of 3 strategies, 4 objective metrics, and 15 classes in the ticketing dataset. The bars in cells depict the improvement in the range $(0\%, 50\%]$, while red texts indicate worsened performance. (b) The results can be summarized as a knowledge map showing the best strategy or strategies for each metric-class combination. For each map region, the strategies within 0.03\% difference from the best strategy are also selected.}
    \label{tab:S3M4tests}
    \vspace{-2mm}
    \hspace{2cm}(a) the results of systematic testing \hspace{3.3cm}
    (b) a summary of the best strategy-metric combinations\\ 
    \includegraphics[width=\linewidth]{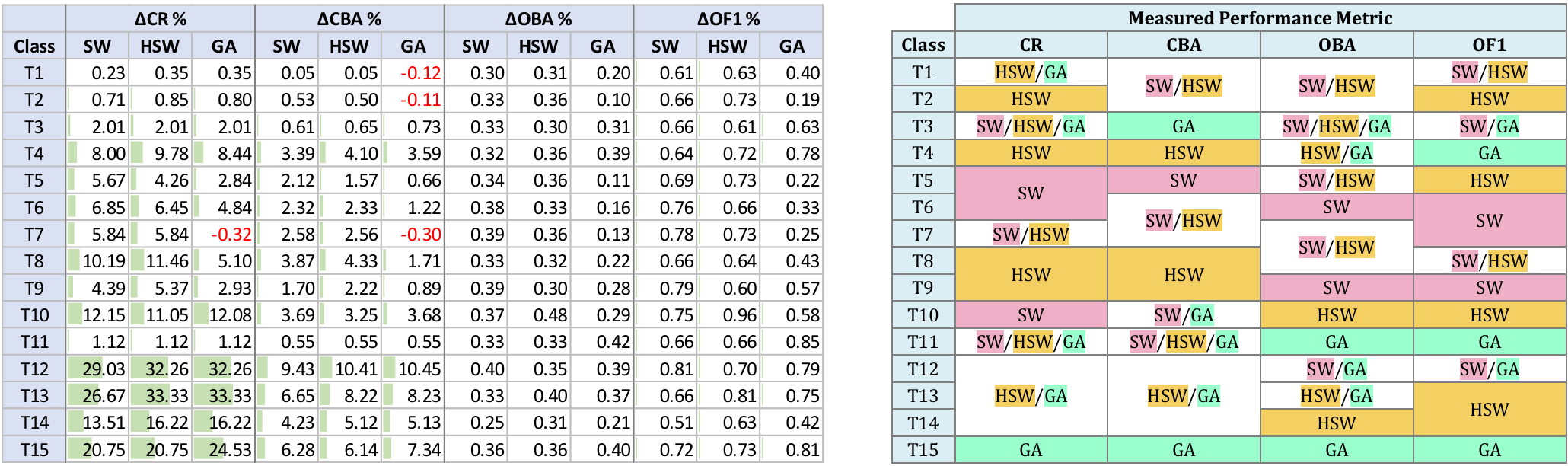}
\end{table*}

\begin{table*}[t]
    \caption{When a model is improved with the SW strategy and objective metric $\phi^\text{cr}$ (class-based recall) for classes $T1, T2, \ldots, T15$ in the ticketing dataset, the direct improvement of the target classes can be seen in the yellow cells on the left part of the table. Meanwhile, there are positive and negative impact on other classes and other performance metrics. The green bars in cells depict the improvement in the range $(0\%, 50\%]$, while red texts indicate negative impact. The last row shows the baseline performance. There are 12 such tables for three strategies (SW, HSW, GA) and four objective metrics ($\phi_\text{cr}, \phi_\text{cba}, \phi_\text{oba}, \phi_\text{of1}$).}
    \label{tab:Impact-SW-CR}
    \vspace{-2mm}
    \includegraphics[width=\linewidth]{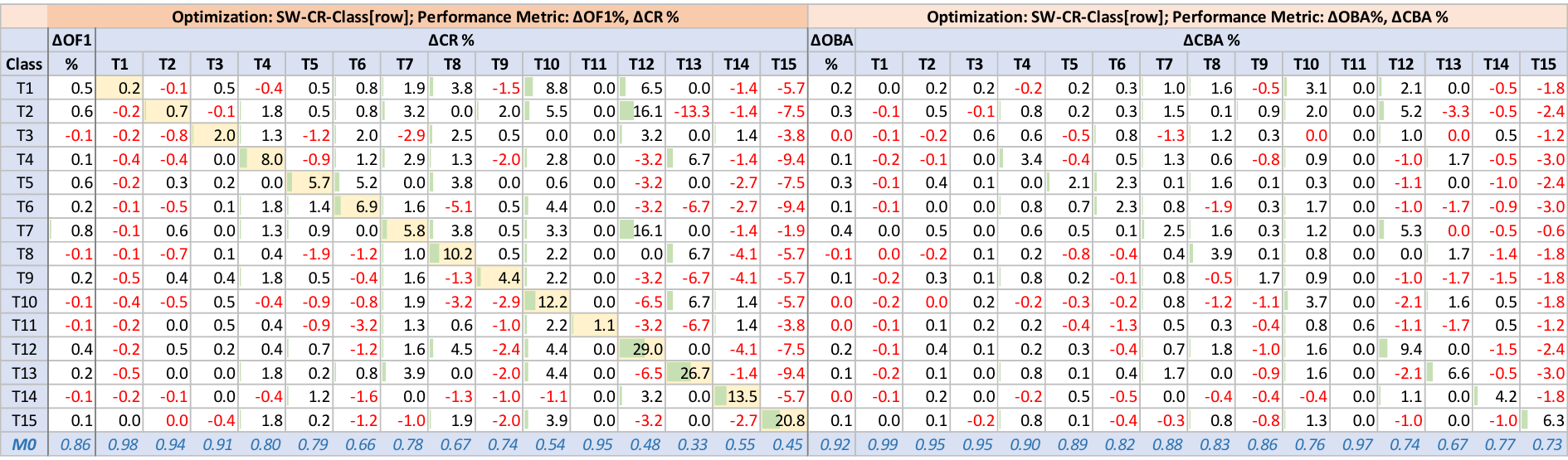}
    \centering
\end{table*}

\vspace{1mm}\noindent
\textbf{Comparison of Strategies and Metrics.}
We systematically tested the combinations of three strategies (SW, HSW, GA) and four objective metrics ($\phi^\text{cr}, \phi^\text{cba}, \phi^\text{oba}, \phi^\text{of1}$). When these combinations were applied to each of the 15 classes where example messages were selected, there were a total of 180 combinations. Table \ref{tab:S3M4tests}(a) shows the results of these 180 combinations. All results were obtained from fixed-time experiments (1 GPU hour).

From the table, we can make several observations:
\begin{itemize}
    \item Class-based objective metrics ($\phi^\text{cr}$ and $\phi^\text{cba}$) mostly delivered improvement for the target classes with a few exceptions. For example, when $\phi^\text{cr}$ is applied to class T1, strategies SW, HSW, and GA improved the recall of T1 by 0.23, 0.35, 0.35 respectively.%
    \item Overall objective metrics ($\phi^\text{oba}$ and $\phi^\text{of1}$) were consistently improved for every class where example messages were selected from.%
    \item Smaller classes, e.g., T12 $\sim$ T15 were noticeably improved.%
    \item Table \ref{tab:S3M4tests}(b) summarizes the strategies that were considered to be the best for each class-metric combination. While each strategy may appear in several neighboring cells that form a color block, there is no overwhelming superior strategy. This observation leads to \textbf{\emph{research question 2}} in Figure \ref{fig:AutoGeTS}.%
\end{itemize}

\noindent
\textbf{Performance Beyond an Objective Metric.}
When a strategy-metric combination was applied to a class in an experiment, we also measure the overall performance of the model and its performance on other classes. Table \ref{tab:Impact-SW-CR} shows such measurements when the combination (SW, $\phi^\text{cr}$) was applied to each of the 15 classes.

For example, in the first row T1, the model retrained with synthestic data generated using examples from the training data of T1 not only improves the recall of T1 (i.e., 0.2\% in the yellow cell), but also improves the two overall measurements $\phi^\text{of1}$ and $\phi^\text{oba}$ by 0.5\% and 0.2\% respectively as well as the class-based measurements $\phi^\text{cr}$ and $\phi^\text{cba}$ for many other classes. Noticeably, $\phi^\text{cr}$ for class T10 was improved by 8.8\%.

From the table, we can also observe that some classes, such as T10, T12, and T13 have often benefited from workflows that were intended to improve other classes. Meanwhile, there were many cells with red numbers indicating worsened performance. Note that the results are representative only for single-phase experiments.

With the 12 combinations of three strategies and four objective metrics, there are 11 other tables similar to Table \ref{tab:Impact-SW-CR}. The results of these experiments can be found in Appendix \ref{apx:Impact}. Hence, if one wishes to improve $\phi^\text{cr}$ for class T1, one can search the column $\Delta$CR\%-T1 in all 12 tables for the best performance, and thereby the best combination strategy-metrix-class. Any of the 180 cell may potentially offer the best setting for improving $\phi^\text{cr}$ for class T1.

The 12 tables of experimental results also inform us that we should not make simple assumption, such as ``\emph{to improve $\phi^\text{cba}$ of class $C$, model improvement using metric $\phi^\text{cba}$ and example messages from class $C$ training data is always the best approach.}'' This leads to a solution for \textbf{\emph{research question 3}} (see Section \ref{sec:Ensemble}). 

\vspace{1mm}\noindent
\textbf{Further Experiments.}
To ensure what observed in the aforementioned experiments described is not an isolated instance, we conducted further experiments using the same experimental environment with different datasets (TREC-6 \cite{li2002learning} and Amazon Reviews 2023 \cite{hou2024bridging}) and an alternative method for generating synthetic data (Easy Data Augmentation (EDA) tool \cite{wei2019eda}). The results of these experiments can be found in Appendix \ref{apx:Validation}.

In general, we can observe a common phenomenon that there is no superior strategy and we cannot assume a superior objective metric. Meanwhile, the amount of improvement through retraining with synthetic data generated using EDA normally is noticeably lower than with synthetic data generated using GPT-3.5.

\section{Ensemble Strategy and Further Experiments}
\label{sec:Ensemble}
The results of our large scale experiments indicated that there was no superior search strategy or objective metric that can address the typical business requirements outlined at the beginning of Section \ref{sec:Methods}. This led us to \textbf{\emph{research question 2}}. We noticed that the results were highly agreeable when we repeated these single-phase experiments.
Although the performance of each combination of
[SW, HSW, GA]%
$\times[\phi^\text{cr},\phi^\text{cba},\phi^\text{oba},\phi^\text{of1}]$
depends many factors (e.g., training data, class size, feature specification, LLMs, and model structure $M^\square$), these factors do not change much in the process for improving a model deployed in a specific ticketing system. Therefore, single-phase experiments (such as those reported in Section \ref{sec:Experiments}) collect knowledge as to what search strategy and what metric may work for each class. In other words, tables (such as the one in Table \ref{tab:Impact-SW-CR}) are essentially quantitative knowledge maps generated by ML training and testing processes. Once these tables become available, the model improvement process can use them to identify more effective strategies, metrics, and classes for a performance metric, and the process can become more cost-effective.

\vspace{1mm}\noindent
\textbf{Multi-Phase Model Improvement.}
Given a model $M_0$ to be improved, the systematic sampling of
$[C_1, C_2, \ldots, C_k]\times$[SW, HSW, GA]%
$\times[\phi^\text{cr},\phi^\text{cba},\phi^\text{oba},\phi^\text{of1}]$
produces a knowledge map. At the end of this process, likely, it also delivers the most improved model $M^*_1$. Because in this systematic sampling phase, each attempt involves example messages from the training data of only one individual class. Naturally, we can initiate a new process for improving $M^*_1$ obtained in the first phase. As exemplified in Table \ref{tab:Impact-SW-CR}, we can also use examples selected from another class to achieve a requirement for improvement.
With 12 such tables, we can also use a different strategy and a different objective metric. Given a single business requirement, we can invoke multiple phases of model improvement with different combinations of strategy, objective metric, and classes for providing example messages.

\vspace{1mm}\noindent
\textbf{Multi-objective Model Improvement.}
Recall the three typical business requirements in Section \ref{sec:Methods}, a model improvement process often has to deal with more than one requirement, e.g., improving the recall of class $X$ and $Y$, while improving the overall accuracy. The experiments reported in Section \ref{sec:Experiments} also indicate such requirements can conflicting with each other if one makes attempts with just one approach (e.g., red numbers in Table \ref{tab:Impact-SW-CR}), while many other combinations of strategy, objective metric, and classes can provide alternative approaches. Once tables similar to Table \ref{tab:Impact-SW-CR}) become available, the model improvement process can use the knowledge stored in these tables to explore different cost-effective approaches to meet multi-objective requirements.

The performance measurements obtained in an early phase can therefore facilitates a knowledge-based algorithm, which provides an answer to \textbf{\emph{research question 3}}.

\begin{figure}[t]
    \centering
    \includegraphics[width=.84\linewidth]{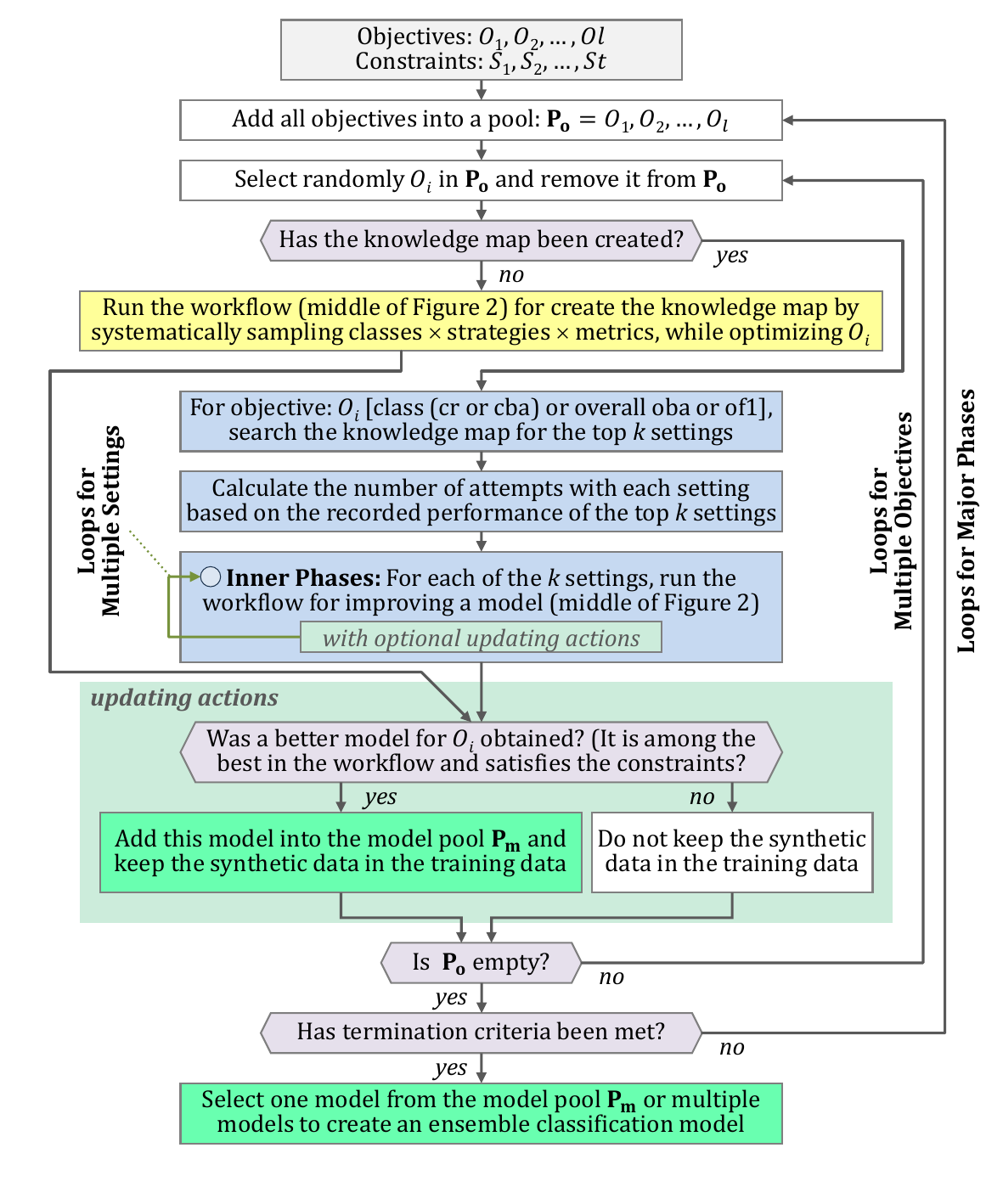}
    \caption{A knowledge-based ensemble algorithm for multi-phase and multi-objective model improvement.}
    \label{fig:EnsembleAlg}
\end{figure}

\begin{figure*}[t]
    \centering
    \includegraphics[width=1\linewidth]{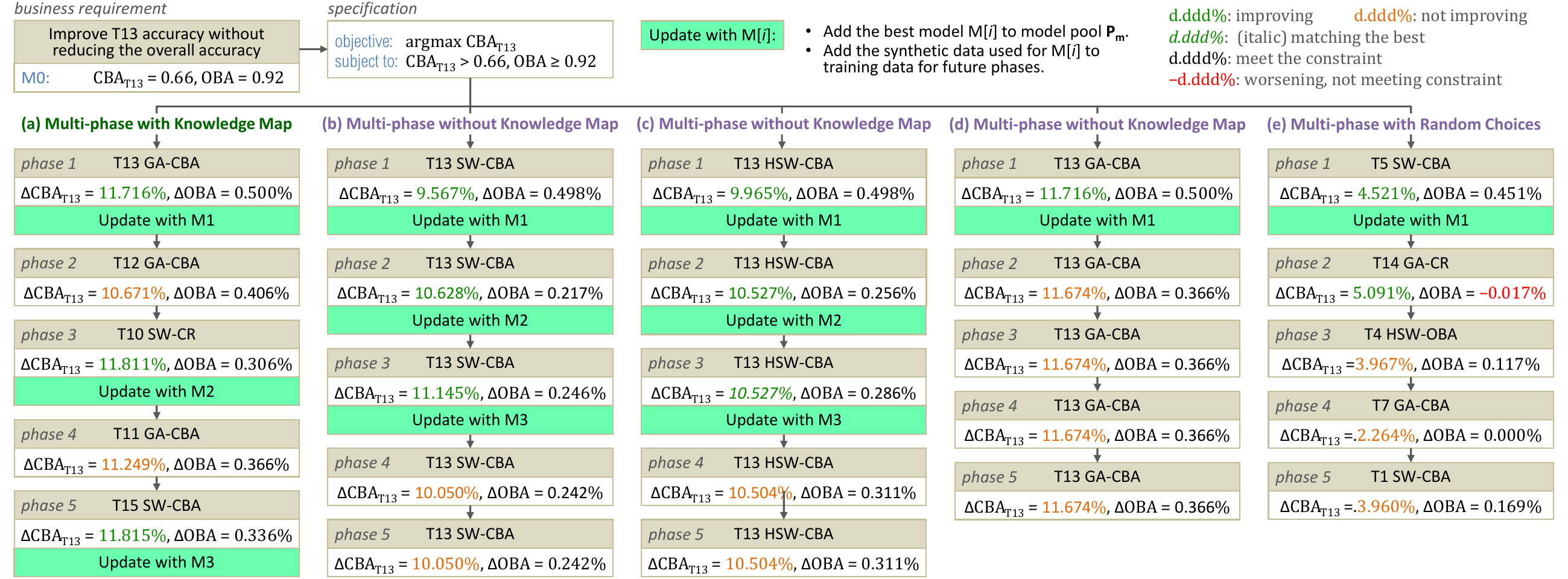}
    \caption{Examples of multi-phase model improvement with and without using a knowledge map. }
    \label{fig:Multi-Phases}
\end{figure*}

\vspace{1mm}\noindent
\textbf{Knowledge-based Ensemble Algorithm.}
Figure \ref{fig:EnsembleAlg} shows the flowchart of an ensemble algorithm for multi-phase and multi-objective model improvement. Given a set of objectives $O_1, O_2, \ldots, O_l$ and a set of constraints $S_1, S_2, \ldots, S_l$, the algorithm begins with the creation of a pool of objectives $\mathbf{P_o}$. The algorithm randomly selects one objective $O_i$ from $\mathbf{P_o}$ each time, aiming to improve the currently-best candidate model with $O_i$ as the focus. The randomness also provides the outer loop of the algorithm with further opportunities to improve the currently-best candidate model.

If the algorithm detects the absence of a knowledge map, it activates the process for creating such a map, while optimizing $O_i$ (i.e., the yellow block). As described in Sections \ref{sec:Methods} and \ref{sec:Experiments}, the process is the middle part of Figure \ref{fig:AutoGeTS} with systematic sampling all combinations of classes, search strategies, and objective metrics. For example, for the ticketing example, the process may run 180 workflows for 15 classes $\times$ 3 strategies $\times$ 4 metrics.

If the knowledge map is already there, the algorithm moves directly to the knowledge-based approach (i.e., the three blue blocks). The algorithm selects top $k>0$ experimental settings that will likely benefit $O_i$ most according to the knowledge map. Note that the selected settings do not have to use the objective metric $\phi$ defined for $O_i$. For example, if $O_i$ is to improve $\phi^{cr}_{C11}$ (the recall of class $C_{11}$), as long as the knowledge map shows that $\phi^{cr}_{C13}$, $\phi^{cba}_{C1}$, and $\phi^{of1}_{C7}$ can also improve the performance measure of $\phi^{cr}_{C11}$, they can be among the candidate objective metrics to be selected.

The algorithm also distributes computational resources according to the knowledge map. Consider the knowledge map shows that the top $k$ settings previously improved $O_i$ by $x_1\%, x_2\%, \ldots, x_k\%$, the algorithm distributes the computational resources proportionally to the $i$-th setting as $(x_i / \sum x_j)\%$ of the total resources available.

\begin{figure*}[t]
    \centering
    \includegraphics[width=\linewidth]{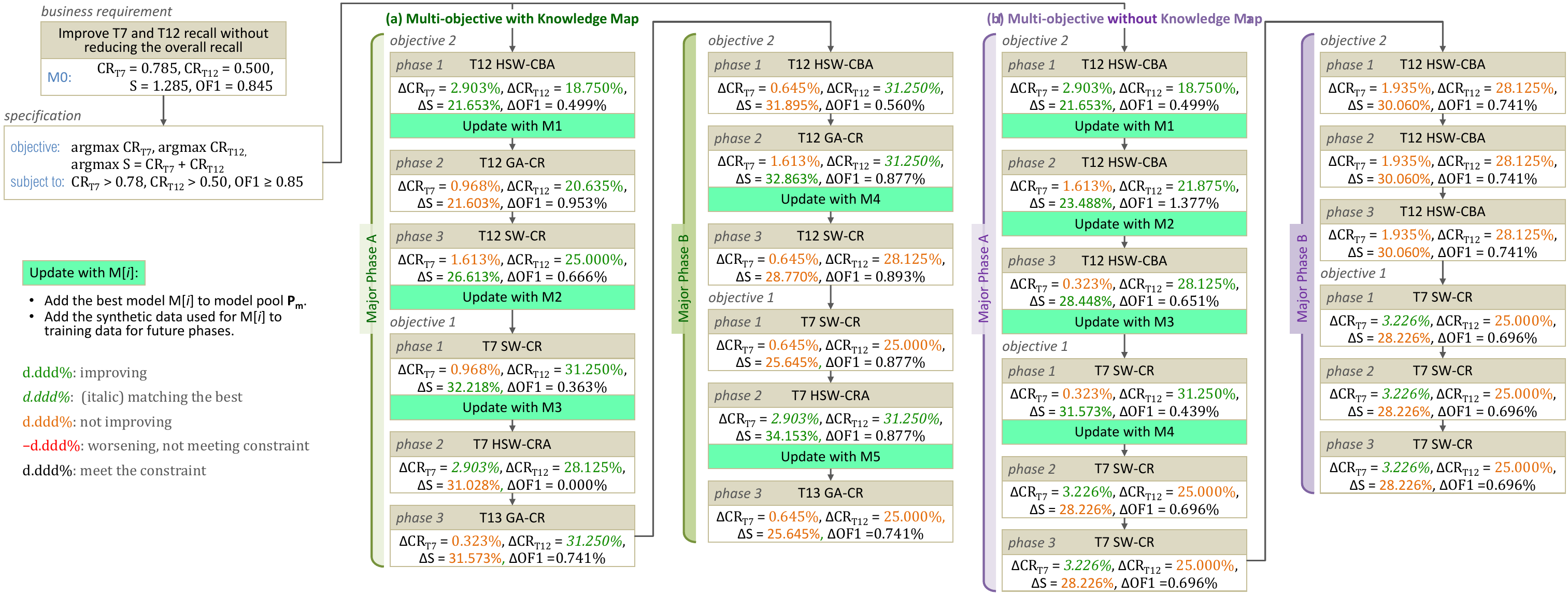}
    \caption{Examples of multi-objective model improvement with and without using a knowledge map.}
    \label{fig:Multi-Objectives}
\end{figure*}

The algorithm adds the best model in each loop to a model pool $\mathbf{P_m}$. At the end of the process, the best model can be selected from $\mathbf{P_m}$ based on a predefined numerical criterion (e.g., a weighted sum of different objective measurement) or by human experts who brought additional operational knowledge into the decision process. Alternatively, one can select a few best models to create an ensemble classification model.

\vspace{1mm}\noindent
\textbf{Further Experiments.}
We conducted further experiments to study the scenarios of multi-phase model improvement for a single objective and for multiple objectives. Figure \ref{fig:Multi-Phases} illustrates the process and results of a set of experiments with a single objective. The business requirement considered class T13 as an important and would like to improve its accuracy. It was first translated to an optimization specification, for which five different processes were invoked, (a) knowledge-based approach, (b,c,d) three brute-force approaches that targeting T13 directly with three different strategies, and (e) a randomly selecting strategy-metric-class combinations. Each approach makes attempts in five phases. For a single-objective, one can select the best model according to the highest performance measurement $\Delta\text{CBA}_\text{T13}$, i.e., 11.815\% resulted from the knowledge-based approach. We conducted similar experiments with different types of single objective, and the knowledge-based approach performed better in the majority of the experiments.     

Figure \ref{fig:Multi-Objectives} illustrates the process and results of a set of experiments with two objectives. The business requirements are to improve the recall of classes T7 and T12. The requirements were first translated to maximization of $\phi^\text{cr}_\text{T7}$ and $\phi^\text{cr}_\text{T12}$, together with a multi-objective assessment function. For this group of experiments, we define a simple function $S = \phi^\text{cr}_\text{T7} + \phi^\text{cr}_\text{T12}$, which will trigger actions of adding a model into the model pool $\mathbf{P_m}$ and adding the corresponding synthetic data to the training data for subsequent phases. In practice, one can define a more complex assessment function and trigger actions, e.g., giving some objectives a high priority, or updating $\mathbf{P_m}$ more frequently than the training data. In the brute-force approach (b) on the right, the first four phases made improvement gradually, and then further improvement became difficult. The knowledge-based approach (a) invoked three different settings for each objective, and made improvement in more distributed manner, reaching the highest $S$ value in the 11$^\text{th}$ inner phase. We conducted similar experiments with different types of multi-objective settings, and the knowledge-based approach performed better in the majority of the experiments.

\section{Conclusions}
\label{sec:Conclusions}
In this work, we developed an automated workflow, AutoGeTS, for improving text classification models, and an approach for using the knowledge discovered in the workflow to guide subsequent model improvement processes. While the work was motivated by challenges in an industrial application, we followed three \textbf{\emph{research questions}}, conducted experiments at a large scale, and validated our findings using public datasets (TREC-6 \cite{li2002learning} and Amazon Reviews 2023 \cite{hou2024bridging}) in addition to the ticketing data, and compared LLMs approach with a traditional data augmentation approach.

In many machine learning workflows, it is common that none of the known individual techniques is superior over others, though most of us are inspired to find the very best solution. A knowledge map is a divide-and-conquer map showing the best solution in each context. Hence a knowledge-based ensemble is piecewise optimization of using the best solution(s) in each context. We envisage that this approach can be adapted in many other practical applications.  

The work also confirmed that using LLMs to generate synthetic data can address a common challenge in the industrial workflow for improving models for ticketing systems deployed in different organization. As LLMs are not part of these classification models, they do not add extra burden on the models. As LLMs improve through version update, they provide sustainable technical support to the AutoGeTS workflow.

In future work, we plan to conduct more large-scale experiments and analyze the results to gain deep understanding of different factors that make some example messages more effective than others, and we hope the use such analysis as useful knowledge for guideline the development of more intelligent and effective techniques to select examples for generating synthetic data.

\newpage

\section*{Acknowledgments}
This work has been made possible by the Network of European Data Scientists. We would like to express our gratitude to the people who facilitated this project, in particular, Dolores Romero Morales from Copenhagen Business School.

\bibliographystyle{unsrt}  
\bibliography{AutoGeTS} 

\appendix

\newpage

\noindent
\begin{center}
\Large\textbf{APPENDICES}\\[1mm]
\large\textbf{AutoGeTS: Knowledge-based Automated Generation of Text Synthetics for Improving Text Classification}\\[3mm]
\end{center}

In the following appendices, we provide further experiment results through visualization plots. The experimental data will be made available on GitHub after the double-blind review process. These appendices include:

\begin{itemize}
    \item Appendix \ref{apx:Validation} provides further experimental results with two additional datasets and with a traditional data augmentation tool. The results confirmed the findings about \textbf{\emph{research question 1}} presented in Section \ref{sec:Experiments}.
    \item Appendix \ref{apx:Impact} provides further experimental results in 11 tables. In total, there are 12 tables, only one table (Table \ref{tab:Impact-SW-CR}) was given in the main body of paper. The results in these 12 tables helped us answer \textbf{\emph{research question 2}} and \textbf{\emph{research question 3}}, and provided a knowledge map for the ensemble algorithm presented in Section \ref{sec:Ensemble}.
\end{itemize}

\section{Validation with the EDA Tool and TREC-6 and Amazon Datasets }
\label{apx:Validation}
In this appendix, we report further experiments for validating observation about the lack of a superior search strategy from the experiments with the ticketing data (Table \ref{tab:S3M4tests}). We conducted these further experiments with two public datasets: TREC-6 \cite{li2002learning} and Amazon Reviews 2023 \cite{hou2024bridging}, in conjunction with the GPT-3.5's API and the Easy Data Augmentation (EDA) tool \cite{wei2019eda}.

\subsection{TREC-6 Dataset with EDA and GPT-3.5}

The TREC-6 dataset \cite{li2002learning} comprises 5542 fact-based questions categorized into six semantic classes with varying class sizes from 86 to 1250 questions. The performance of the original CatBoost classification model trained on TREC-6 (without synthetic data) is shown in Table~\ref{tab:M0 Model_TREC-6}.

\begin{table}[H]
\caption{The performance of the original CatBoost model $M_0$ trained on the TREC-6 dataset (without synthetic data).}
\label{tab:M0 Model_TREC-6}
\centering
\renewcommand{\arraystretch}{1}
\begin{tabular}{@{}|c|c|c|c|c|@{}}
    \hline
     Class & Class Size & Balanced Accuracy & Recall & F1-Score \\
    \hline
    ENTY & 1250 & 0.861 & 0.825 & 0.757 \\
    HUM & 1223 & 0.903 & 0.850 & 0.846 \\
    DESC & 1162 & 0.881 & 0.802 & 0.820 \\
    NUM & 896 & 0.908 & 0.836 & 0.866 \\
    LOC & 835 & 0.882 & 0.789 & 0.819 \\
    ABBR & 86 & 0.761 & 0.522 & 0.686 \\
    \hline
    Overall & 5542 & 0.889 & 0.816 & 0.816 \\
    \hline
\end{tabular}
\end{table}

Similar to the ticketing data, we used both the EDA tool and the GPT-3.5 to generate synthetic data, and ran experiments with all combinations of three strategies, four objective metrics, and six classes where example questions were selected as prompts. The EDA results were shown in Table \ref{tab:TREC6-EDA} and the GPT 3.5 results were shown in Table \ref{tab:TREC6-LLM}.

\begin{table*}[ht]
    \centering
    \caption{Systematic experimentation with the TREC-6 dataset and synthetic data generated by the EDA tool.
    (a) The results of systematic testing of 72 combinations of 3 strategies, 4 objective metrics, and 6 classes in the ticketing dataset. The bars in cells depict the improvement in the range $(0\%, 50\%]$, while red texts indicate worsened performance. (b) The results can be summarized as a knowledge map showing the best strategy or strategies for each metric-class combination. For each map region, the strategies within 0.03\% difference from the best strategy are also selected.}
    \label{tab:TREC6-EDA}
    \vspace{-2mm}
    \hspace{2cm}(a) the results of systematic testing \hspace{3.3cm}
    (b) a summary of the best strategy-metric combinations\\
    \includegraphics[width=\linewidth]{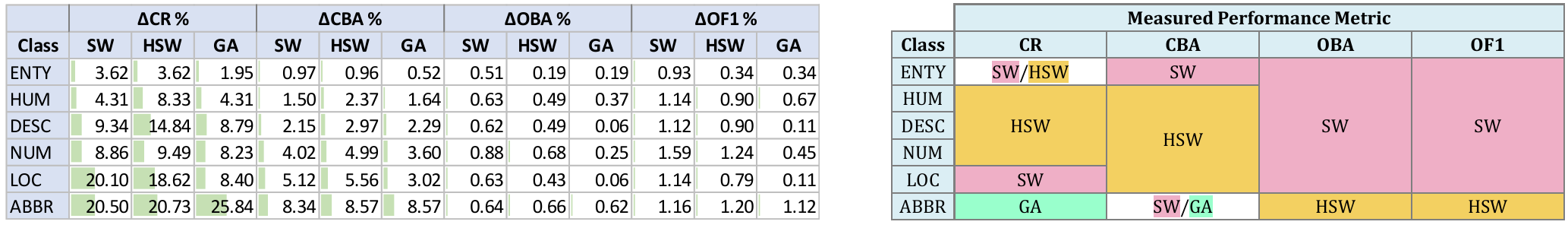}
\end{table*}

\begin{table*}[t]
    \centering
    \caption{Systematic experimentation with the TREC-6 dataset and synthetic data generated by the GPT-3.5 API.
    (a) The results of systematic testing of 72 combinations of 3 strategies, 4 objective metrics, and 6 classes in the ticketing dataset. The bars in cells depict the improvement in the range $(0\%, 50\%]$, while red texts indicate worsened performance.
    (b) The results can be summarized as a knowledge map showing the best strategy or strategies for each metric-class combination. For each map region, the strategies within 0.03\% difference from the best strategy are also selected.}
    \label{tab:TREC6-LLM}
    \vspace{-2mm}
    \hspace{2cm}(a) the results of systematic testing \hspace{3.3cm}
    (b) a summary of the best strategy-metric combinations\\
    \includegraphics[width=\linewidth]{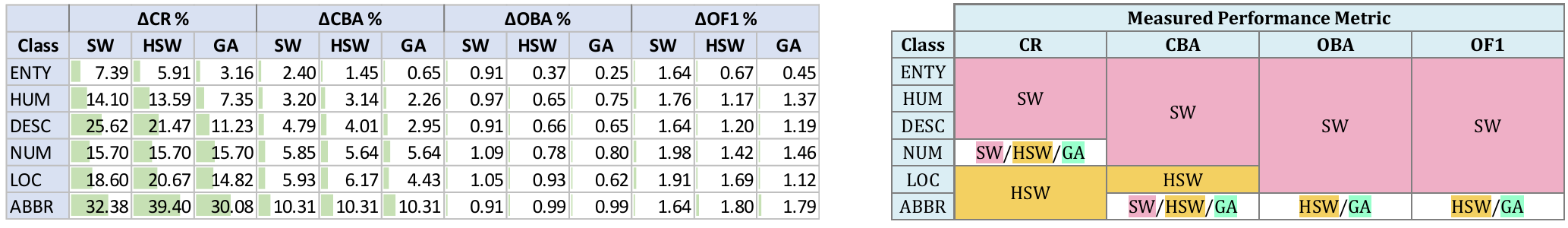}
\end{table*}

From these two tables, we can observe:
\begin{itemize}
    \item The amount of improvement in Table \ref{tab:TREC6-LLM} is in general higher than that in Table \ref{tab:TREC6-EDA}, confirming the advantage of using LLMs to generate synthetic data.
    \item With the synthetic data generated using the EDA tool, the strategy HSW worked better for several classes when classed-measures were used as the objective measures. The strategy SW worked better in most cases when overall measures were used as the objective measures.
    \item With the synthetic data generated using the GPT 3.5 API, the strategy SW worked better in most cases when across the whole table. However, there are five cells where SW is not the preferred strategy.
    \item Both tables confirmed that the knowledge-based approach can enable the selection of more effective strategies for individual classes and objective metrics.   
\end{itemize}

\subsection{Amazon Subset with EDA and GPT-3.5}

With the Amazon Reviews’2023 dataset \cite{hou2024bridging}, a subset of 10000 reviews were randomly selected from the Gift Cards subject. The subset has five rating classes with different class sizes ranging from 121 to 8389 reviews. The performance of the original CatBoost classification model trained on this Amazon Reviews subset (without synthetic data) is shown in Table~\ref{tab:M0 Model_Amazon_Review}.

\begin{table}[H]
\caption{The performance of the original CatBoost model $M_0$ trained on the Amazon Review subset (without synthetic data).}
\label{tab:M0 Model_Amazon_Review}
\centering
\renewcommand{\arraystretch}{1}
\begin{tabular}{@{}|c|c|c|c|c|@{}}
    \hline
     Class & Class Size & Balanced Accuracy & Recall & F1-Score \\
    \hline
    R1 & 807 & 0.834 & 0.705 & 0.650 \\
    R2 & 121 & 0.499 & 0 & 0 \\
    R3 & 206 & 0.524 & 0.049 & 0.089 \\
    R4 & 477 & 0.586 & 0.174 & 0.288 \\
    R5 & 8389 & 0.758 & 0.977 & 0.950 \\
    \hline
    Overall & 10000 & 0.934 & 0.894 & 0.894 \\
    \hline
\end{tabular}
\end{table}

We can notice that for class R2, there was no true positive results, and the recall and F1-score are both 0. For R3 and R4, the measure of the recall and F1-score measures are very low. Because these are small classes in comparison with R5, misleadingly, the overall measures appear fairly reasonable. 

Similar to the ticketing data and TREC-6 data, we used both the EDA tool and the GPT-3.5 to generate synthetic data, and ran experiments with all combinations of three strategies, four objective metrics, and five classes where example reviews were selected as prompts. The direct testing results of the experiments with the EDA tool were shown in Table \ref{tab:EDA_Based_Results_AmazonReview} and the direct testing results with the GPT 3.5 API were shown in Table \ref{tab:LLM_Based_Results_AmazonReview}. Note that in both tables, the recall values (CR) for class R2 are now above zero. Clearly synthetic data helped.

\begin{table*}[t]
\centering
\renewcommand{\arraystretch}{1}
\caption{Amazon Reviews Subset EDA-Based Results}
\label{tab:EDA_Based_Results_AmazonReview}
\begin{tabular}{@{}c|ccc|ccc|ccc|ccc@{}}
\hline
 & \multicolumn{3}{c|}{CR} & \multicolumn{3}{c|}{CBA} & \multicolumn{3}{c|}{OBA} & \multicolumn{3}{c}{OF1}\\
\cline{2-13}
Class & SW & HSW & GA & SW & HSW & GA & SW & HSW & GA & SW & HSW & GA \\
\hline
R1 & 0.732 & 0.763 & 0.732 & 0.845 & 0.852 & 0.846 & 0.936 & 0.936 & 0.938 & 0.898 & 0.898 & 0.900 \\
R2 & 0.067 & 0.053 & 0.067 & 0.526 & 0.526 & 0.526 & 0.938 & 0.938 & 0.938 & 0.901 & 0.900 & 0.900 \\
R3 & 0.122 & 0.098 & 0.098 & 0.560 & 0.560 & 0.559 & 0.937 & 0.937 & 0.937 & 0.900 & 0.899 & 0.899 \\
R4 & 0.221 & 0.233 & 0.233 & 0.601 & 0.603 & 0.603 & 0.936 & 0.938 & 0.936 & 0.898 & 0.900 & 0.898 \\
R5 & 0.984 & 0.984 & 0.984 & 0.782 & 0.784 & 0.784 & 0.938 & 0.938 & 0.938 & 0.900 & 0.901 & 0.901 \\
\hline
\end{tabular}
\end{table*}
\begin{table*}[t]
\centering
\renewcommand{\arraystretch}{1}
\caption{Amazon Reviews Subset AEDA-Based Results}
\label{tab:AEDA_Based_Results_AmazonReview}
\begin{tabular}{@{}c|ccc|ccc|ccc|ccc@{}}
\hline
 & \multicolumn{3}{c|}{CR} & \multicolumn{3}{c|}{CBA} & \multicolumn{3}{c|}{OBA} & \multicolumn{3}{c}{OF1}\\
\cline{2-13}
Class & SW & HSW & GA & SW & HSW & GA & SW & HSW & GA & SW & HSW & GA \\
\hline
R1 & 0.739 & 0.788 & 0.651 & 0.852 & 0.853 & 0.813 & 0.938 & 0.938 & 0.937 & 0.901 & 0.901 & 0.899 \\
R2 & 0.105 & 0.067 & 0.158 & 0.578 & 0.578 & 0.552 & 0.938 & 0.938 & 0.937 & 0.900 & 0.900 & 0.900 \\
R3 & 0.122 & 0.188 & 0.098 & 0.560 & 0.549 & 0.548 & 0.938 & 0.937 & 0.938 & 0.900 & 0.900 & 0.901 \\
R4 & 0.233 & 0.348 & 0.197 & 0.615 & 0.614 & 0.597 & 0.937 & 0.938 & 0.938 & 0.900 & 0.901 & 0.900 \\
R5 & 0.984 & 0.987 & 0.975 & 0.811 & 0.812 & 0.825 & 0.938 & 0.938 & 0.934 & 0.901 & 0.901 & 0.895 \\
\hline
\end{tabular}
\end{table*}
\begin{table*}[t]
\centering
\renewcommand{\arraystretch}{1}
\caption{Amazon Reviews Subset LLM-Based Results}
\label{tab:LLM_Based_Results_AmazonReview}
\begin{tabular}{@{}c|ccc|ccc|ccc|ccc@{}}
\hline
 & \multicolumn{3}{c|}{CR} & \multicolumn{3}{c|}{CBA} & \multicolumn{3}{c|}{OBA} & \multicolumn{3}{c}{OF1}\\
\cline{2-13}
Class & SW & HSW & GA & SW & HSW & GA & SW & HSW & GA & SW & HSW & GA \\
\hline
R1 & 0.745 & 0.797 & 0.745 & 0.855 & 0.866 & 0.854 & 0.938 & 0.938 & 0.938 & 0.901 & 0.902 & 0.901 \\
R2 & 0.158 & 0.158 & 0.158 & 0.579 & 0.579 & 0.579 & 0.940 & 0.939 & 0.938 & 0.904 & 0.902 & 0.901 \\
R3 & 0.146 & 0.188 & 0.146 & 0.573 & 0.570 & 0.571 & 0.938 & 0.938 & 0.938 & 0.901 & 0.902 & 0.901 \\
R4 & 0.348 & 0.348 & 0.326 & 0.615 & 0.614 & 0.608 & 0.938 & 0.938 & 0.938 & 0.901 & 0.901 & 0.901 \\
R5 & 0.989 & 0.997 & 0.995 & 0.828 & 0.822 & 0.825 & 0.938 & 0.939 & 0.940 & 0.902 & 0.903 & 0.905 \\
\hline
\end{tabular}
\end{table*}

Similarly we showed the amount of improvement in Table \ref{tab:Amazon-EDA} and \ref{tab:Amazon-LLM} respectively. From these two tables, we can observe:
\begin{itemize}
    \item The amount of improvement in Table \ref{tab:Amazon-LLM} is in general higher than that in Table \ref{tab:Amazon-EDA}, confirming the advantage of using LLMs to generate synthetic data.
    \item Using synthetic data resulted in noticeable improvement on the recall measures of classes R2, R3, and R4.
    \item There was no obvious superior strategy.
    \item Both tables confirmed that the knowledge-based approach can enable the selection of more effective strategies for individual classes and objective metrics.   
\end{itemize}

\begin{table*}[t]
    \centering
    \caption{Systematic experimentation with a subset of the Amazon Review dataset and synthetic data generated by the EDA tool.
    (a) The results of systematic testing of 60 combinations of 3 strategies, 4 objective metrics, and 5 classes in the ticketing dataset. The bars in cells depict the improvement in the range $(0\%, 50\%]$, while red texts indicate worsened performance.
    For the six dark green cells on the left, the original recall measures were zero or near zero. Hence the improvement is well over 50\%. The $\Delta$CR\% values are shown as white text over dark green background. \#\#\#\#\#\# indicates infinity. 
    (b) The results can be summarized as a knowledge map showing the best strategy or strategies for each metric-class combination. For each map region, the strategies within 0.03\% difference from the best strategy are also selected.}
    \label{tab:Amazon-EDA}
    \vspace{-2mm}
    \hspace{2cm}(a) the results of systematic testing \hspace{3.3cm}
    (b) a summary of the best strategy-metric combinations\\
    \includegraphics[width=\linewidth]{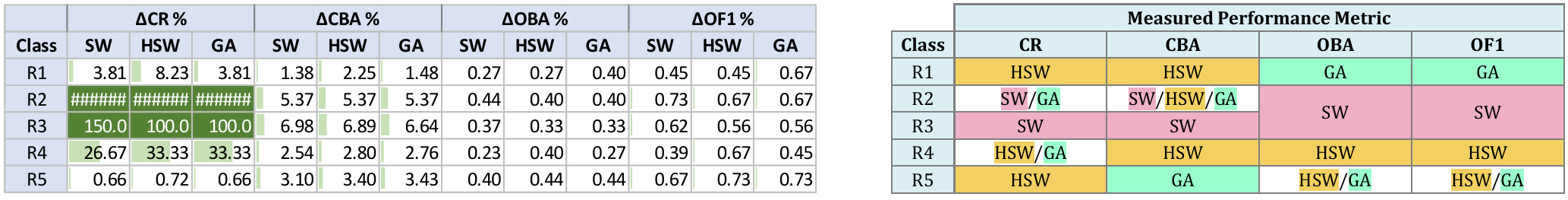}
\end{table*}

\begin{table*}[t]
    \centering
    \caption{Systematic experimentation with a subset of the Amazon Review dataset and synthetic data generated by the GPT-3.5 API.
    (a) The results of systematic testing of 60 combinations of 3 strategies, 4 objective metrics, and 5 classes in the ticketing dataset. The bars in cells depict the improvement in the range $(0\%, 50\%]$, while red texts indicate worsened performance.
    For the nine dark green cells on the left, the original recall measures were 0 or near zero. Hence the improvement is well over 50\%. The $\Delta$CR\% values are shown as white text over dark green background. \#\#\#\#\#\# indicates infinity.
    (b) The results can be summarized as a knowledge map showing the best strategy or strategies for each metric-class combination. For each map region, the strategies within 0.03\% difference from the best strategy are also selected.}
    \label{tab:Amazon-LLM}
    \vspace{-2mm}
    \hspace{2cm}(a) the results of systematic testing \hspace{3.3cm}
    (b) a summary of the best strategy-metric combinations\\
    \includegraphics[width=\linewidth]{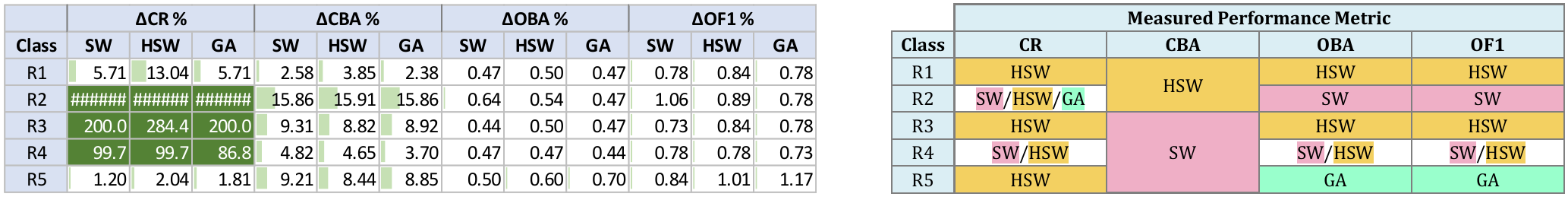}
\end{table*}
\newpage
~
\newpage

\section{Impact on Other Performance Metrics}
\label{apx:Impact}

As mentioned in Section \ref{sec:Experiments}, with the ticketing data, there are 180 combinations of three strategies, four objective metrics, and 15 classes where examples were selected from their training data as the prompts for generating synthetic data. Given each of these combinations, in addition to the objective metric on the specific class, we also measure other performance metrics, including two overall metrics (overall balanced accuracy (OBA) and overall F1-score (OF1)) as well as two class-based metrics (recall (CR) and balanced accuracy (CBA)) on all classes. Table \ref{tab:Impact-SW-CR} showed the cross-impact results of 15 of the 180 combinations, i.e., SW strategy, $\phi^\text{cr}$ (class-based recall), and 15 classes. In this appendix, we provide the the cross-impact results for the other $15 \times 11$ combinations in Tables \ref{tab:Impact-HSW-CR} $\sim$ \ref{tab:Impact-GA-OF1}.

In addition, from these tables, we can also observe many interesting phenomena, for example:

\begin{itemize}
    \item Most targeted classes, i.e., the yellow cells in these tables benefited from the improvement process with positive values (black numbers) in these cell. Red numbers in yellow cells are in general uncommon.
    \item The recall measure of T12 is often improved noticeable even when it was not the target class, e.g., in Tables \ref{tab:Impact-SW-CR}, \ref{tab:Impact-SW-CBA}, \ref{tab:Impact-HSW-CBA}, \ref{tab:Impact-SW-OBA}, \ref{tab:Impact-HSW-OBA}, \ref{tab:Impact-GA-OBA}, \ref{tab:Impact-SW-OF1}, \ref{tab:Impact-HSW-OF1}, and \ref{tab:Impact-GA-OF1}, where the green bars in the column $\Delta$CR\% T12 are noticeable.

\end{itemize}

\textbf{Most importantly, the data in such tables provides the ensemble algorithm in Section \ref{sec:Ensemble} with the knowledge map.}

\begin{table*}[t]
    \caption{Similar to Table \ref{tab:Impact-SW-CR}, when a model is improved with the HSW strategy and objective metric $\phi^\text{cr}$ (class-based recall) for classes $T1, T2, \ldots, T15$ in the ticketing dataset, the direct improvement of the target classes can be seen in the yellow cells on the left part of the table. Meanwhile, there are positive and negative impact on other classes and other performance metrics. The green bars in cells depict the improvement in the range $(0\%, 50\%]$, while red texts indicate negative impact. The last row shows the baseline performance.}
    \label{tab:Impact-HSW-CR}
    \vspace{-2mm}
    \includegraphics[width=\linewidth]{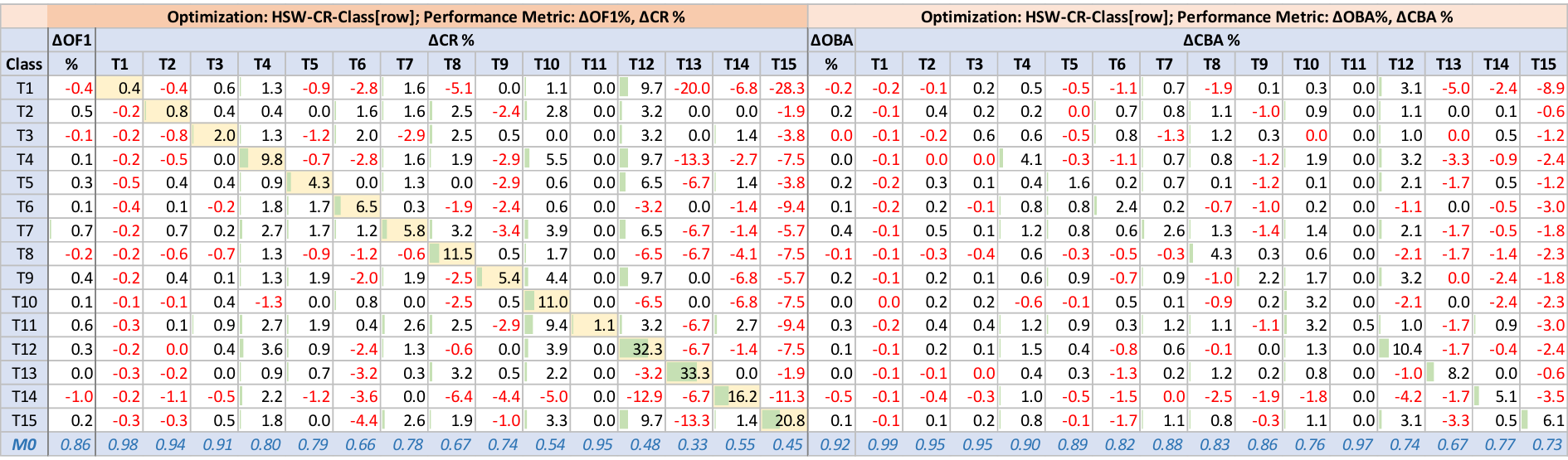}
    \centering
\end{table*}

\begin{table*}[t]
    \caption{Similar to Table \ref{tab:Impact-SW-CR} and Table \ref{tab:Impact-HSW-CR}, this table shows the impact of the GA strategy and objective metric $\phi^\text{cr}$ (class-based recall) for classes $T1, T2, \ldots, T15$ in the ticketing dataset.}
    \label{tab:Impact-GA-CR}
    \vspace{-2mm}
    \includegraphics[width=\linewidth]{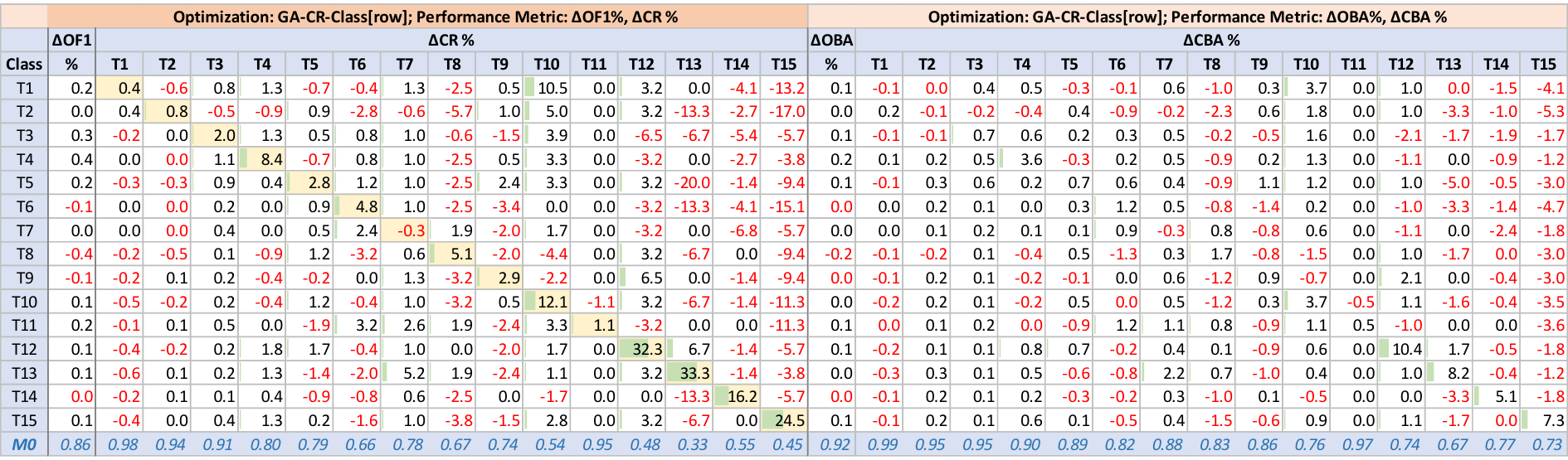}
    \centering
\end{table*}

\begin{table*}[t]
    \caption{Similar to Table \ref{tab:Impact-SW-CR} and Tables \ref{tab:Impact-HSW-CR}$\sim$\ref{tab:Impact-GA-CR}, this table shows the impact of the SW strategy and objective metric $\phi^\text{cba}$ (class-based balanced accuracy) for classes $T1, T2, \ldots, T15$ in the ticketing dataset.}
    \label{tab:Impact-SW-CBA}
    \vspace{-2mm}
    \includegraphics[width=\linewidth]{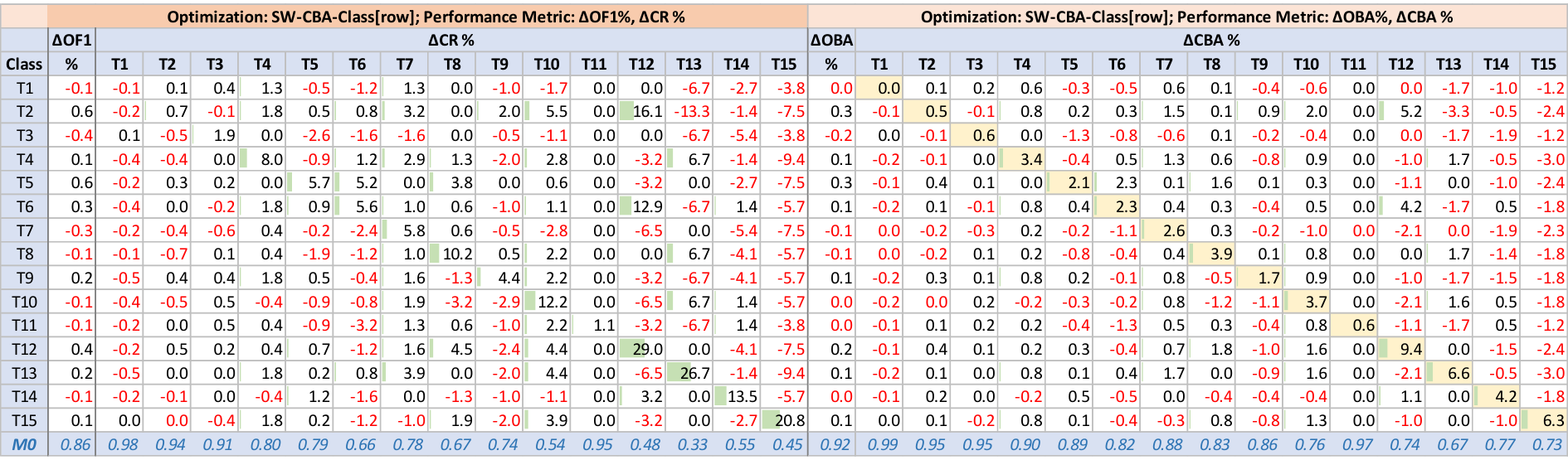}
    \centering
\end{table*}

\begin{table*}[t]
    \caption{Similar to Table \ref{tab:Impact-SW-CR} and Tables \ref{tab:Impact-HSW-CR}$\sim$\ref{tab:Impact-SW-CBA}, this table shows the impact of the HSW strategy and objective metric $\phi^\text{cba}$ (class-based balanced accuracy) for classes $T1, T2, \ldots, T15$ in the ticketing dataset.}
    \label{tab:Impact-HSW-CBA}
    \vspace{-2mm}
    \includegraphics[width=\linewidth]{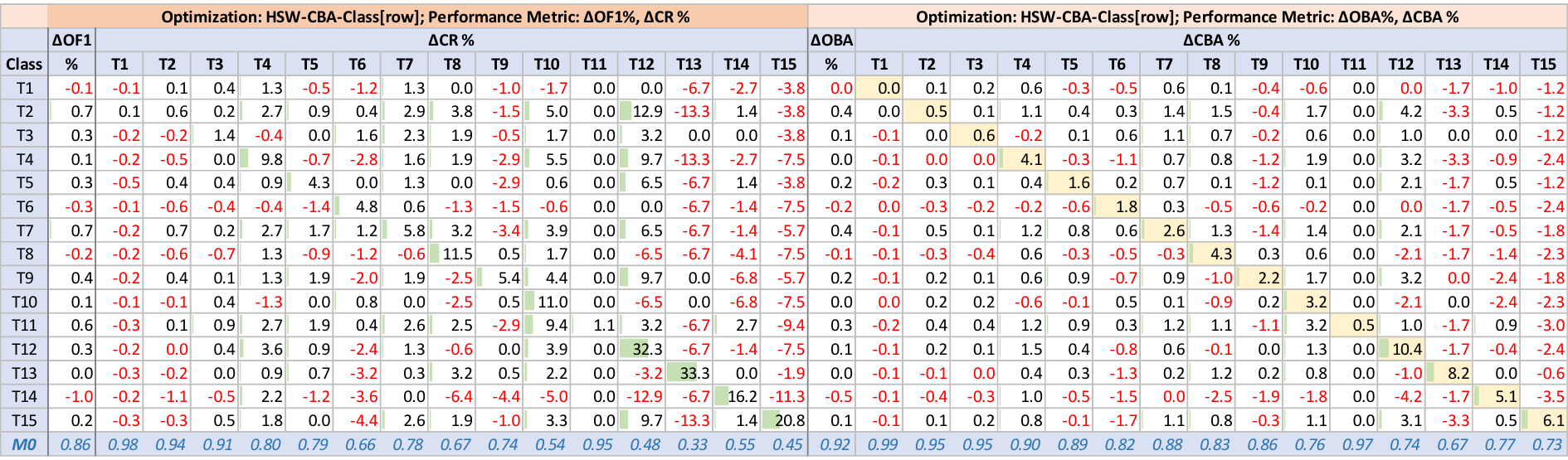}
    \centering
\end{table*}

\begin{table*}[t]
    \caption{Similar to Table \ref{tab:Impact-SW-CR} and Tables \ref{tab:Impact-HSW-CR}$\sim$\ref{tab:Impact-HSW-CBA}, this table shows the impact of the GA strategy and objective metric $\phi^\text{cba}$ (class-based balanced accuracy) for classes $T1, T2, \ldots, T15$ in the ticketing dataset.}
    \label{tab:Impact-GA-CBA}
    \vspace{-2mm}
    \includegraphics[width=\linewidth]{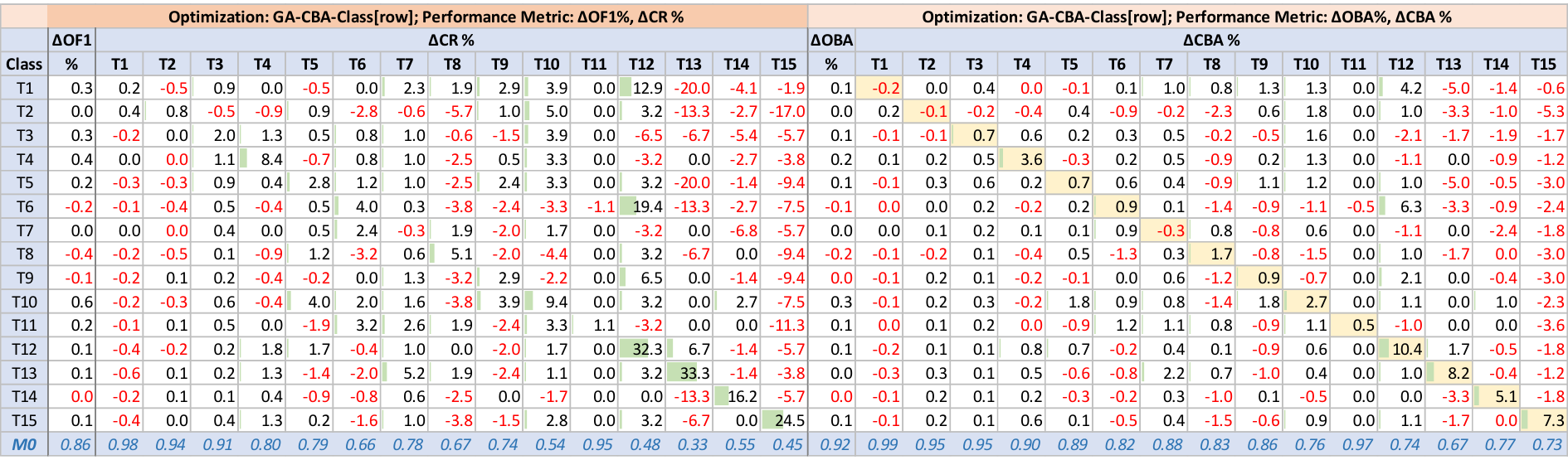}
    \centering
\end{table*}

\begin{table*}[t]
    \caption{Similar to Table \ref{tab:Impact-SW-CR} and Tables \ref{tab:Impact-HSW-CR}$\sim$\ref{tab:Impact-GA-CBA}, this table shows the impact of the SW strategy and objective metric $\phi^\text{oba}$ (overall balanced accuracy) for classes $T1, T2, \ldots, T15$ in the ticketing dataset.}
    \label{tab:Impact-SW-OBA}
    \vspace{-2mm}
    \includegraphics[width=\linewidth]{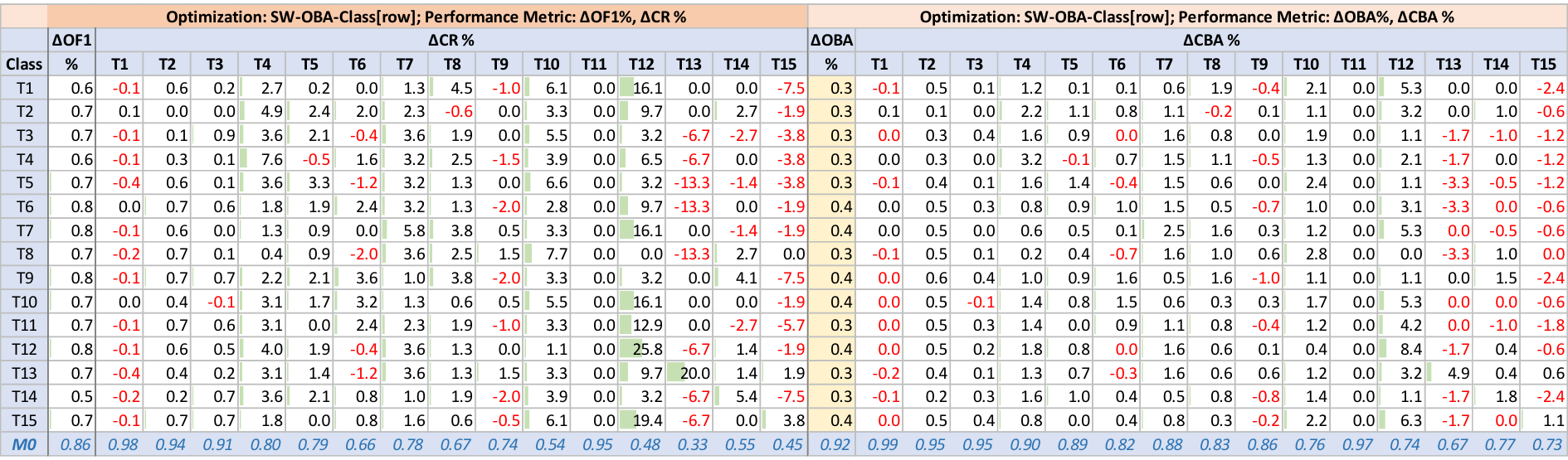}
    \centering
\end{table*}

\begin{table*}[t]
    \caption{Similar to Table \ref{tab:Impact-SW-CR} and Tables \ref{tab:Impact-HSW-CR}$\sim$\ref{tab:Impact-SW-OBA}, this table shows the impact of the HSW strategy and objective metric $\phi^\text{oba}$ (overall balanced accuracy) for classes $T1, T2, \ldots, T15$ in the ticketing dataset.}
    \label{tab:Impact-HSW-OBA}
    \vspace{-2mm}
    \includegraphics[width=\linewidth]{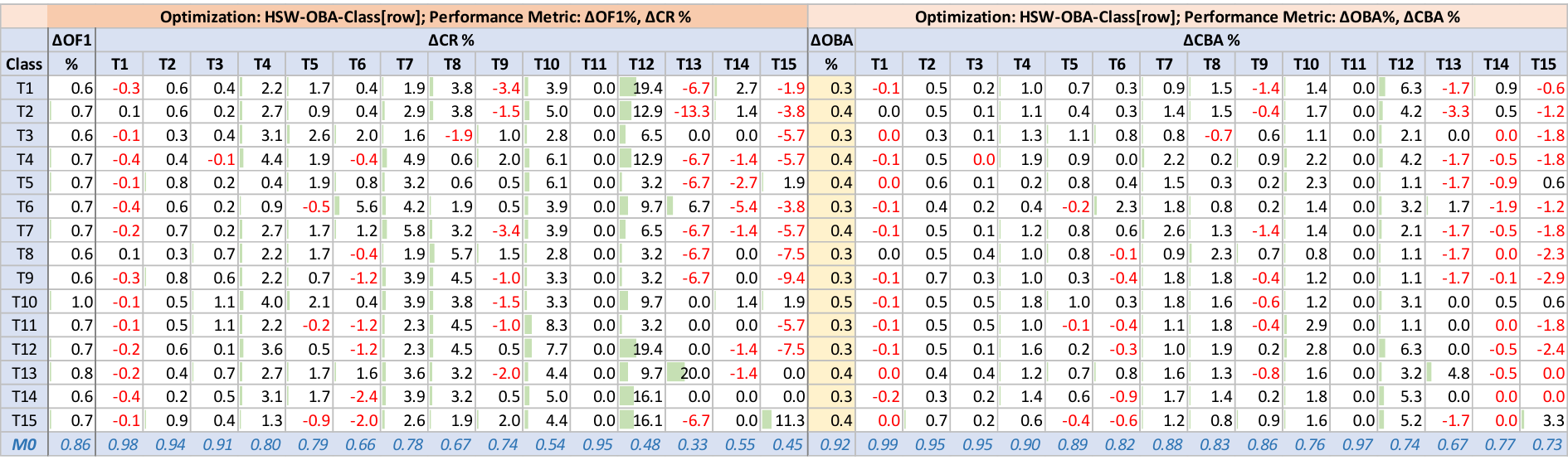}
    \centering
\end{table*}

\begin{table*}[t]
    \caption{Similar to Table \ref{tab:Impact-SW-CR} and Tables \ref{tab:Impact-HSW-CR}$\sim$\ref{tab:Impact-HSW-OBA}, this table shows the impact of the GA strategy and objective metric $\phi^\text{oba}$ (overall balanced accuracy) for classes $T1, T2, \ldots, T15$ in the ticketing dataset.}
    \label{tab:Impact-GA-OBA}
    \vspace{-2mm}
    \includegraphics[width=\linewidth]{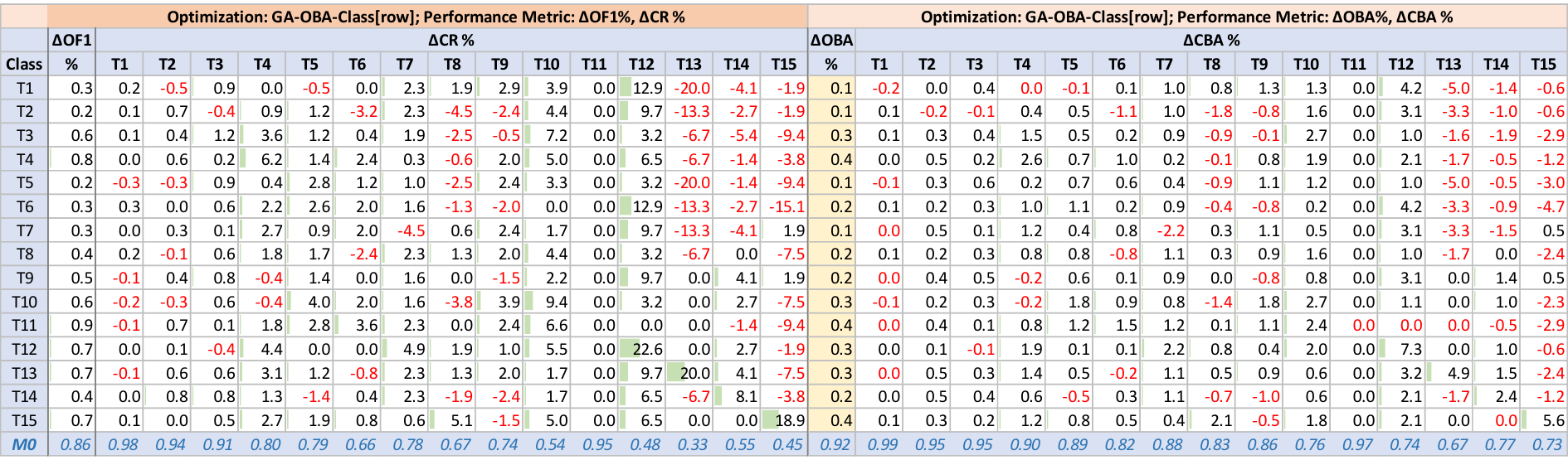}
    \centering
\end{table*}

\begin{table*}[t]
    \caption{Similar to Table \ref{tab:Impact-SW-CR} and Tables \ref{tab:Impact-HSW-CR}$\sim$\ref{tab:Impact-GA-OBA}, this table shows the impact of the SW strategy and objective metric $\phi^\text{of1}$ (overall F1-score) for classes $T1, T2, \ldots, T15$ in the ticketing dataset.}
    \label{tab:Impact-SW-OF1}
    \vspace{-2mm}
    \includegraphics[width=\linewidth]{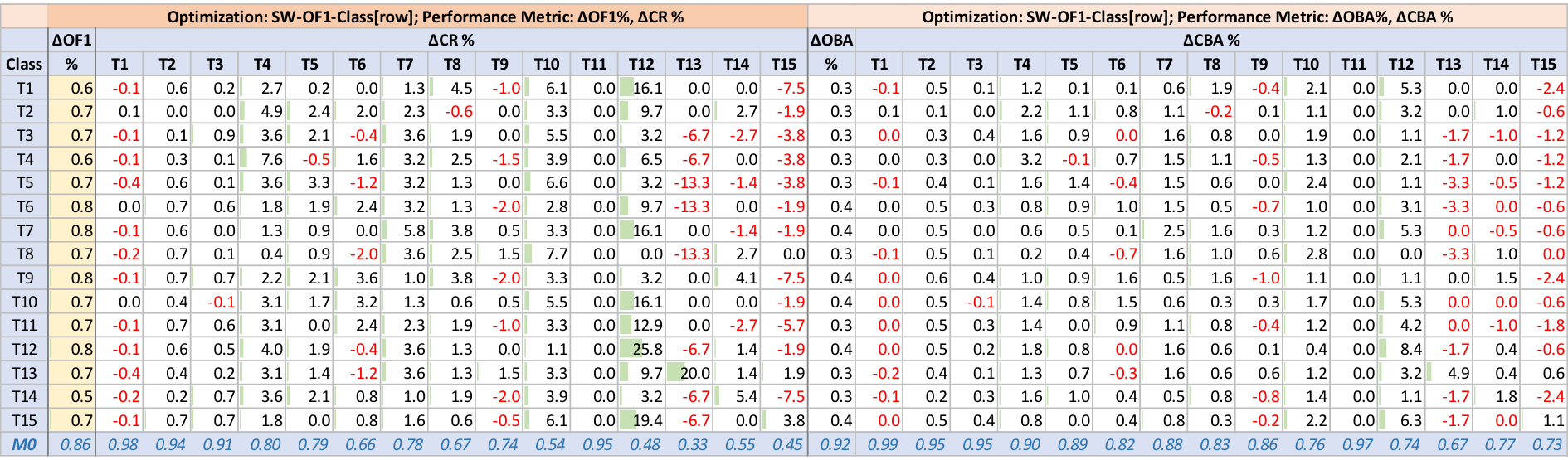}
    \centering
\end{table*}

\begin{table*}[t]
    \caption{Similar to Table \ref{tab:Impact-SW-CR} and Tables \ref{tab:Impact-HSW-CR}$\sim$\ref{tab:Impact-SW-OF1}, this table shows the impact of the HSW strategy and objective metric $\phi^\text{of1}$ (overall F1-score) for classes $T1, T2, \ldots, T15$ in the ticketing dataset.}
    \label{tab:Impact-HSW-OF1}
    \vspace{-2mm}
    \includegraphics[width=\linewidth]{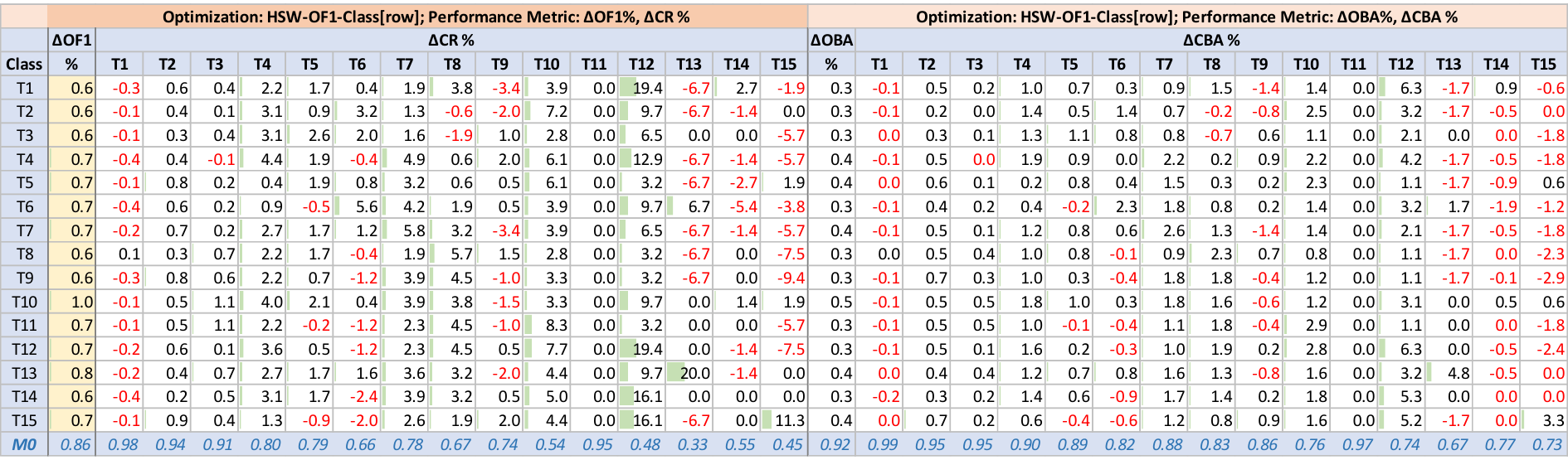}
    \centering
\end{table*}

\begin{table*}[t]
    \caption{Similar to Table \ref{tab:Impact-SW-CR} and Tables \ref{tab:Impact-HSW-CR}$\sim$\ref{tab:Impact-HSW-OF1}, this table shows the impact of the GA strategy and objective metric $\phi^\text{of1}$ (overall F1-score) for classes $T1, T2, \ldots, T15$ in the ticketing dataset.}
    \label{tab:Impact-GA-OF1}
    \vspace{-2mm}
    \includegraphics[width=\linewidth]{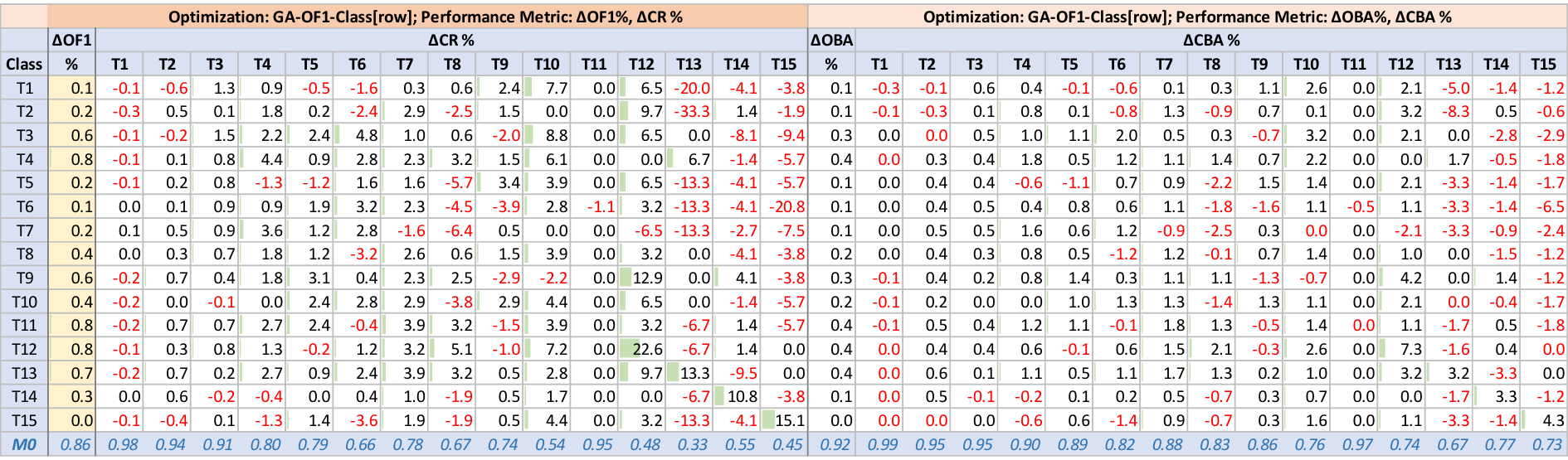}
    \centering
\end{table*}

\end{document}